\documentclass[11pt]{article}
\usepackage[preprint]{acl}

\usepackage{amsmath}
\usepackage{amssymb}
\usepackage{times}
\usepackage{latexsym}
\usepackage{booktabs}
\usepackage{array}
\usepackage{tabularx}
\usepackage{placeins}
\usepackage{capt-of}

\usepackage[T1]{fontenc}
\usepackage[utf8]{inputenc}
\usepackage{microtype}
\usepackage{inconsolata}
\usepackage{graphicx}
\usepackage{xurl}

\usepackage{xcolor}
\usepackage{tikz}
\usepackage{pgfplots}
\usepgfplotslibrary{groupplots}
\usetikzlibrary{arrows.meta}
\pgfplotsset{compat=1.18}

\definecolor{localblue}{HTML}{4C78A8}
\definecolor{plangenorange}{HTML}{F58518}
\definecolor{interventionred}{HTML}{E45756}
\definecolor{interventiondark}{HTML}{C7372F}
\definecolor{gridgray}{HTML}{D0D0D0}
\definecolor{gategreen}{HTML}{3F8F3F}

\title{Evaluating Compressed Memory:\\ Dependence, Utility, and Content Benefit}

\author{Minsu Oh$^{1}$ \quad Seunghui Jwa$^{1}$ \quad Chanjun Park$^{2,*}$ \quad Yeo-Chan Yoon$^{1}$\thanks{Corresponding authors: Yeo-Chan Yoon and Chanjun Park.} \\
$^{1}$Department of Artificial Intelligence, Jeju National University \\
$^{2}$School of Software, Soongsil University \\
\texttt{ycyoon@jejunu.ac.kr}}

\hypersetup{pdftitle={Evaluating Compressed Memory: Dependence, Utility, and Content Benefit},pdfauthor={Minsu Oh, Seunghui Jwa, Chanjun Park, Yeo-Chan Yoon}}
\begin{document}
\maketitle

\begin{abstract}
Does a language model answer better because its compressed memory contains useful document information? Replacing memory vectors with zeros can lower a trained model's score without showing that it outperforms a model trained without memory. We combine zeroing with separately trained models that cannot access stored text or receive the same constant input for every document. Across nine systems, some models rely on memory yet perform worse than the trained alternative. For example, in experiments that predict states in generated stories, zeroing memory worsens a model's prediction loss, yet a model trained without memory achieves lower loss than the original model with intact memory. On TriviaQA, giving xRAG memory from another question's passage raises F1, which measures word overlap with accepted answers, from 19.38 without a document to 24.78. Its effect on how often outputs contain an accepted answer is statistically uncertain. The question's retrieved passage scores higher still (29.88 F1). Thus a gain over no document alone does not establish that the gain requires the question's passage. Evaluations must therefore specify the comparison model, absolute performance, and generation and scoring conditions.

\end{abstract}

\section{Introduction}
Compressed memory encodes text as vectors that a large language model (LLM) can use in place of the original text. Evaluating memory therefore requires evidence that the stored information improves prediction. A lower score after removing memory, or a higher score than a model given no document, can appear to provide that evidence. {We show that neither comparison alone establishes a benefit from the stored content.} A model can rely on memory yet predict worse than a model trained without it; a module trained to supply the same input for every document can also improve scores.

The problem is a mismatch between the comparison performed and the conclusion drawn. Removing memory after training asks whether the model relies on an input it learned to use. It does not ask whether training with memory produced a better predictor than training without it. Even that second comparison leaves another question: does the advantage require document content, or can a module trained without access to that content also improve performance? Treating these questions as interchangeable leaves the value of stored information untested.

\begin{figure*}[!t]
\centering
\includegraphics[width=0.84\textwidth,trim=4pt 6pt 4pt 6pt,clip]{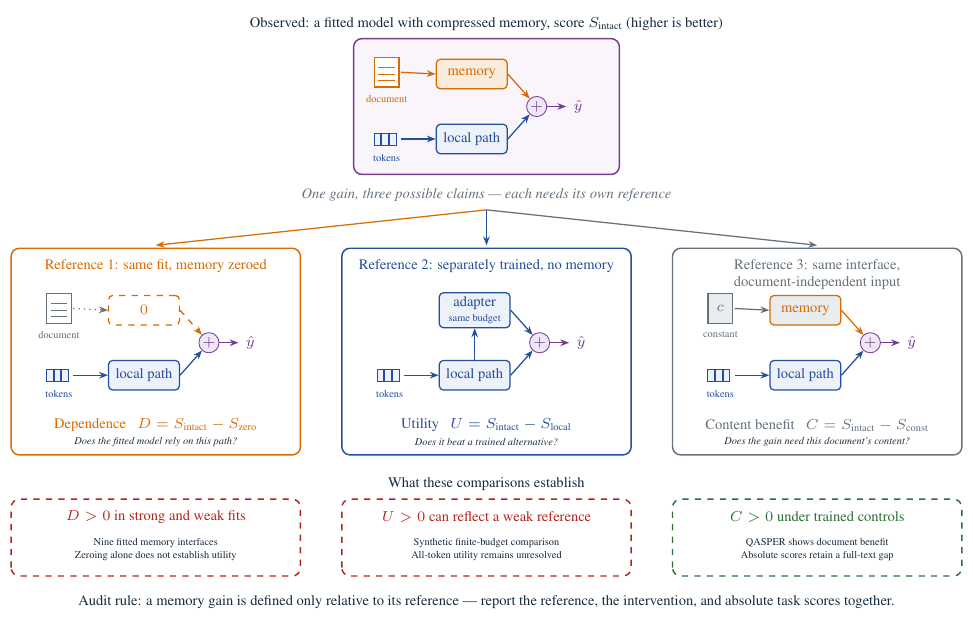}
\caption{Three comparisons for interpreting a memory gain. Zeroing memory tests whether an already trained model relies on it. Separately trained models without memory or with constant input test whether memory and document content improve performance.}
\label{fig:memory-concept}
\end{figure*}

These distinctions build on established comparisons with retrained models and controls for learned representations \citep{hooker2019benchmark,hewitt2019control}. We bring the comparisons together to test what memory evaluations establish (Figure~\ref{fig:memory-concept}). We measure {dependence} by zeroing memory in an already trained model, and {utility} by comparing against a model trained without the tested memory. To measure {content benefit}, we train the same memory module with document vectors or with one constant input shared by every document. The constant-input module can learn to help the LLM, but cannot convey the current document's facts. Our contribution is an empirical account of where these comparisons lead to different conclusions and which additional evidence supports a claim about document content.

\begin{enumerate}
\item {We demonstrate that zeroing damage does not establish a training benefit.} Across nine systems, four are harmed by zeroing memory yet perform worse than their trained alternatives on generated-story state prediction. These cases have low task performance; they do not establish the same disagreement in models that solve the task well. A separate controlled study produces a positive memory advantage when both models worsen, because the model without memory worsens more. Together, these results show why zeroing damage and a relative gain must be checked against trained alternatives and absolute performance (Section~\ref{sec:controlled}).

\item {We show that zeroing damage can include a benefit learned without document content.} On questions about scientific papers, zeroing the output of a module trained on constant input lowers F1 by 5.73 points. F1 measures word overlap with an accepted answer, accounting for missing and extra words. This drop occurs although the module receives no facts from the current document. Training on document vectors also outperforms training on constant input, providing separate evidence that document content helps. These results distinguish the benefit of a learned input from the additional benefit of document content in a working question-answering system (Section~\ref{sec:gold-qa}).

\item {We establish that a gain over no document need not be specific to the question's retrieved passage.} Using xRAG, which maps a document embedding to one LLM input vector \citep{cheng2024xrag}, we find that passages from other questions improve some scores over no document on three datasets. The question's retrieved passage performs better. Comparing only against no document would miss that some improvement also occurs with a different passage. We also show that generation settings and scoring rules change the conclusions supported by these comparisons (Section~\ref{sec:native-generation}).
\end{enumerate}

These findings make memory evaluation a question of matching evidence to a claim: reliance requires an intervention, utility requires a trained comparison, and document-content benefit requires a comparison that withholds document information. Reporting them together makes clear what a memory gain does, and does not, demonstrate.

\section{Related Work}
\subsection{Interpreting Performance Gains}
\citet{oconnor2021context} compare changing evaluation context with training on altered contexts. Our memory audit applies this distinction by pairing zeroing after training with a separately trained model without memory. Conditional probing tests what a model's internal text vectors add to a prediction after accounting for information already available to the predictor \citep{hewitt2021conditional}. For example, do contextual word vectors help predict a word's part of speech beyond knowing which word it is? Our constant-input comparison asks a related question: does document input improve performance beyond a trained input that contains no document-specific information?

The controls are established. Probe control tasks help distinguish representation properties from a probe's ability to learn labels \citep{hewitt2019control}. Remove and Retrain (ROAR) tests feature importance by retraining after removal \citep{hooker2019benchmark}. Learned prompts supply trainable inputs without document content \citep{li2021prefix,lester2021prompt}, and optimal ablation learns a constant replacement \citep{li2024optimalablation}. Our constant-input projector produces one learned vector. Activation patching replaces internal model values with values computed from a different input to test which components affect a prediction. \citet{zhang2024patching} show that how the input is altered and which evaluation metric is used can change which components appear important. Our memory audit likewise distinguishes zeroing from replacement with another document's vectors and reports the scoring rule for each comparison.

We study compressed memory by bringing these controls together in state prediction, word prediction, and question answering. Our contribution is to test whether conclusions about memory change when reliance on an already trained input is compared with the advantage over a separately trained model. We also compare modules trained with document vectors or constant input to determine whether a gain requires document content. This connects established control methods to the specific claims made about compressed memory: that a model uses it, benefits from training with it, or benefits from its stored content.

\subsection{Evaluating Compressed Memory}
Trained comparison models provide precedents for our distinction between memory dependence and utility. Gist, which compresses task instructions, compares models trained with full or hidden instructions \citep[Section~4.2]{mu2023gist}. AutoCompressors, which summarize earlier text into vectors, use fine-tuned language models as baselines \citep[Section~4.1]{chevalier2023autocompressors}. We build on such comparisons by testing whether damage from zeroing memory agrees with the advantage over a model trained without it. To examine this question across memory designs, our experiments include the In-context Autoencoder (ICAE) \citep{ge2024icae}, Compressed Context Memory (CCM) \citep{kim2024compressed}, Activation Beacon \citep{zhang2025beacon}, and MemoryLLM \citep{wang2024memoryllm}, which retain earlier text in compressed form. Each is evaluated against its stated trained alternative (Section~\ref{sec:controlled}).

Our question-answering (QA) studies separate gains from additional training from gains that require document content. xRAG includes a low-rank adaptation (LoRA) control for additional training data \citep[Section~6.2]{cheng2024xrag}; LoRA trains low-rank weight updates \citep{hu2022lora}. COCOM trains a model without documents and studies training the answer-generating LLM, or \emph{decoder}, together with the compressor \citep{rau2025cocom}. LLoCO trains decoder LoRA while keeping compressed document memory fixed \citep{tan2024lloco}. Our decoder study keeps document memory fixed and uses equal training budgets for models trained on correct documents, other documents, or no documents. Both models trained with document memory receive the same correct-document memory at evaluation. Their difference therefore tests the benefit of learning from correct document content under the stated training procedure (Section~\ref{sec:gold-qa}).

Information-preservation studies ask what can be recovered from compression: reconstruction \citep{kuratov2025capacity}, grounding and recall \citep{lajewska-etal-2025-understanding,deng-etal-2025-silver}, or answers lost after compression \citep{belikova2026overflow}. Our question is whether the stored content improves prediction beyond a model trained without that content. CCM already examines learning that bypasses context during compression training \citep{kim2024compressed}. We test for gains of this kind by training a memory module with one constant input shared by all documents. Comparing it with the module trained on document vectors separates a gain achievable without current-document facts from an additional gain with those facts.

The choice of evaluation input and score also affects what a memory gain establishes. Random passages can improve retrieval-augmented generation \citep{cuconasu2024noise}, so an advantage over no document alone need not be specific to the retrieved passage. We examine this issue in compressed memory by comparing the question's retrieved passage, another question's passage, and no document, then testing whether conclusions change with generation settings and scoring rules (Section~\ref{sec:native-generation}). BenchPress compares compressed scores with reference scores \citep{feldman2026nomean}. Our analysis of scores normalized relative to models with no document or with evidence text shows why such normalization cannot by itself establish document use: a model trained on constant input can also receive a positive normalized score (Appendix~\ref{app:reference-audit}).

\section{Comparisons for Memory Evaluation}
\label{sec:method}
\subsection{Three Questions, Three Comparisons}
An \emph{intact} model receives its usual compressed memory. We evaluate all conditions on the same items. A \emph{reference} is the named comparison condition or model, not necessarily with the same training history.

\paragraph{Dependence: intervene after training.}
We zero memory values with model weights fixed. A performance drop shows dependence on those values. The model did not train without memory, so this test cannot establish superiority to one that was.

\paragraph{Utility: train without memory.}
Comparison model $r$ learns the same task without the tested memory. A \emph{local adapter} adds trainable computation without memory access. A \emph{closed-book} QA model receives the question without a document. A \emph{suffix-only} model receives recent tokens without earlier text. We call the memory model's advantage over the specified model its utility, $U_r$ (or $U$). It depends on both models' training and selection.

\paragraph{Content benefit: train with constant input.}
The \emph{constant-input} model receives the same input, computed from training data, for every document. We train the same memory module in both conditions, but constant input cannot convey the current document's facts. The document model's advantage is content benefit, $C$. Equal architectures need not have equal input-dependent capabilities; this difference does not isolate a unique fraction of the gain caused by information.

Negative log-likelihood (NLL) measures how little probability a model assigns to the correct next token. Lower is better; we use nats. The differences are
\begin{align}
D &= \mathrm{NLL}_{\rm zero}-\mathrm{NLL}_{\rm intact},\\
U_r &= \mathrm{NLL}_{r}-\mathrm{NLL}_{\rm intact},\\
C &= \mathrm{NLL}_{\rm constant}-\mathrm{NLL}_{\rm intact}.
\end{align}
F1 measures word overlap between a generated answer and an accepted answer, accounting for both missing and extra words. For a higher-is-better score $S$, such as F1 or accuracy, they are $D=S_{\rm intact}-S_{\rm zero}$, $U_r=S_{\rm intact}-S_r$, and $C=S_{\rm intact}-S_{\rm constant}$. Positive values favor intact memory over the named reference. Thus $D>0$ and $U<0$ mean zeroing hurts, yet a trained alternative does better.

\subsection{What Is Held Fixed}
Our PlanGen test model compares a memory module with a trainable local adapter that uses only recent text. The memory module summarizes consecutive token groups into memory vectors, available after a delay. The LLM combines recent-token states with either the mean of available vectors (\emph{pooled} memory) or one selected vector (\emph{indexed} memory). A local adapter matches the parameter budget but receives no memory vectors (Appendix~\ref{sec:plangen}).

The projector-training studies for QA and word prediction use xRAG's single memory vector. The SFR-Embedding-Mistral encoder (SFR) maps text to a fixed-length vector \citep{salesforce2024sfr}. A trainable \emph{projector} transforms this vector into an LLM input. Each condition trains its own copy from the same projector weights, keeping the encoder and LLM frozen. The projector receives document content, locally visible text, or a constant. This preserves its architecture, but not its information access or inference cost.

\paragraph{A shared zero-input predictor.}
In this frozen-LLM interface, zeroing projector output leaves the same LLM and prompt in every condition. Let their common zero-input score be $S_0$, and write $D_{\rm const}=S_{\rm const}-S_0$. Then
\begin{equation}
D_{\rm doc}=S_{\rm doc}-S_0=C+D_{\rm const}.
\label{eq:common-zero}
\end{equation}
Thus the performance lost by zeroing document memory includes the benefit of a learned constant input. The identity holds for NLL with reversed signs and paired averages. It does not provide another test of document use. In PlanGen, separately trained parameters, including LoRA weights, remain after zeroing, so predictors can differ. Likewise, $D-U=S_{\rm local}-S_{\rm zero}$ compares the trained local model with the zeroed model; it does not separate causal mechanisms.
\label{eq:diagnostic-gap}

\subsection{Content Checks and Uncertainty}
\label{sec:reporting}
A \emph{memory swap} replaces a document's memory with another document's memory. \emph{Evidence masking} hides text that supports the answer; \emph{sham masking} hides equally long text at a similar position outside the designated evidence. These tests change content after training (Appendices~\ref{sec:plangen} and~\ref{app:external-du}). Their effects depend on the intervention: zeroing PlanGen memory retains projection bias, and reordering vectors leaves their pooled mean unchanged.

We report each model's absolute score alongside its advantage over the named comparison. In the decoder-training study, we keep the precomputed document memory fixed and train LLMs with correct documents, other documents, or no documents.

Uncertainty intervals retain the pairing between conditions. Earlier analyses average training seeds per item and resample documents or books while holding the trained models fixed. \emph{Crossed} intervals additionally resample training-order seeds. Generation analyses resample passage groups, joining groups that share a generation batch. Unless otherwise stated, intervals use bootstrap percentiles 2.5 and 97.5. \emph{Simultaneous} intervals adjust for multiple comparisons in a specified family. An interval above zero supports a positive difference; an interval containing zero does not establish absence of an effect. Numerical benchmarks describe magnitudes, not validated thresholds for equivalence (Appendix~\ref{app:statistics}).

\section{Dependence and Utility}
\label{sec:controlled}
\paragraph{Comparing nine systems.}
Does damage from zeroing memory identify models that benefit from training with memory? Table~\ref{tab:main-systems} compares zeroing damage, $D$, with the advantage over a trained model without that memory, $U$. The tasks predict state values assigned earlier in generated stories and words repeated in the PG-19 book corpus \citep{rae2020compressive}. The comparison models receive only recent text. Alongside the compressors described above, we test StreamingLLM \citep{xiao2023streamingllm} and H$_2$O \citep{zhang2023h2o}, which retain selected attention key/value (KV) states from earlier tokens. We also test the convolutional compressor LoCoCo \citep{cai2024lococo} and M+, an extension of MemoryLLM \citep{wang2025mplus}. Each system uses its stated adaptation procedure (Appendix~\ref{app:nine-config}).

On generated stories, CCM, ICAE, Beacon, M+ and MemoryLLM have low prediction losses: 0.029--0.078 NLL, compared with 1.872--2.244 for their local baselines. CCM, Beacon and MemoryLLM also outperform their baselines on PG-19. Four other systems are harmed by zeroing on the synthetic task yet perform worse than their trained baselines. These comparisons concern specific training procedures; they do not rank architectures.

\begin{table*}[t]
\centering\fontsize{8.5}{9.5}\selectfont
\renewcommand{\arraystretch}{0.97}
\setlength{\tabcolsep}{4pt}
\begin{tabular}{@{}llrrrr@{}}
\toprule
System & Task adaptation / reference & Synthetic $U$ & Synthetic $D$ & PG-19 $U$ & PG-19 $D$ \\
\midrule
StreamingLLM \citep{xiao2023streamingllm} & KV / restricted decoder & $-0.610$ & $+6.484$ & $+0.333$ & $+4.895$ \\
H$_2$O \citep{zhang2023h2o} & KV / restricted decoder & $-0.219$ & $+7.690$ & $+4.560$ & $+8.726$ \\
xRAG \citep{cheng2024xrag} & projector / suffix decoder & $-0.932$ & $+6.790$ & $-0.100$ & $+1.237$ \\
LoCoCo \citep{cai2024lococo} & convolution / suffix adapter & $-1.414$ & $+6.367$ & $+0.676$ & $+5.653$ \\
CCM \citep{kim2024compressed} & merge adapter / suffix adapter & $+2.153$ & $+9.910$ & $+0.462$ & $+2.200$ \\
ICAE \citep{ge2024icae} & decoder / suffix decoder & $+2.185$ & $+7.767$ & $-2.932$ & $-2.326$ \\
Beacon \citep{zhang2025beacon} & decoder / suffix decoder & $+1.865$ & $+7.955$ & $+2.861$ & $+4.650$ \\
MemoryLLM \citep{wang2024memoryllm} & decoder / suffix decoder & $+1.847$ & $+9.822$ & $+2.513$ & $+8.106$ \\
M+ \citep{wang2025mplus} & decoder / suffix decoder & $+1.804$ & $+8.834$ & $-3.583$ & $+1.782$ \\
\bottomrule
\end{tabular}
\caption{Advantage over a trained restricted-input model ($U$) and zeroing damage ($D$), in nats; three seeds per row. Positive values favor intact memory. Study-specific simultaneous 95\% intervals exclude zero for every sign (Table~\ref{tab:core-external-bounds}). Synthetic KV budgets: 24 updates with memory versus $183+64+512$ for the restricted-input model. KV policies share weights trained with full context; the last four compressors remain frozen during adaptation. Access and optimization differ across rows (Appendix~\ref{app:external-du}).}
\label{tab:main-systems}
\end{table*}

On generated stories, hiding answer-supporting text hurts H$_2$O and xRAG more than hiding matched non-evidence text. The differences are $+0.405\;[0.353,0.460]$ and $+1.268\;[1.216,1.321]$ nats under their respective simultaneous intervals. Yet H$_2$O has NLL 2.068 versus its baseline's 1.849, and xRAG has 3.211 versus 2.279. Sensitivity to relevant text therefore does not guarantee an advantage over a separately trained model. These interventions differ from training a constant-input model to measure $C$. This xRAG result concerns state prediction and a suffix-only baseline; the positive QA result below uses a different task and question-input baseline.

\subsection{Synthetic Boundary Cases}
\label{sec:training}\label{sec:attention-control}
PlanGen restricts attention to recent text, so only memory can provide the earlier state assignment. Given preceding gold story text, it predicts the token distinguishing the required state value. We use six Pythia and Llama base models \citep{biderman2023pythia,grattafiori2024llama,meta2024llama32}, four conditions, and three seeds, giving 72 trained models. All six pooled-memory configurations have $D>0,U<0$; for Llama-3.2-3B, $D=0.4643$ and $U=-0.1417$ nats. Yet accuracy on 512 rechecked targets is only 14.78\% with prediction restricted to candidate values or over the full vocabulary. Predicting from training frequencies gives 18.55\%. This disagreement occurs in a model that predicts these states unreliably. A further comparison with shared initial weights and equal development-search opportunities does not establish both $D>0$ and $U<0$ (Appendix~\ref{app:training}).

A separate study asks whether concentrating training on the answer improves this result. The local adapter, pooled memory, and constant-input model share ten paired training orders and equal development-search budgets. \emph{All-token} training computes loss throughout the story; \emph{target-only} training computes it only at designated answer tokens, with the same preceding story as input. Held-out answer-token NLL rises from 2.319 to 3.273 for the adapter, 2.319 to 2.492 for pooled memory, and 2.311 to 2.457 for constant input (Table~\ref{tab:objective-absolute-main}).

\begin{table}[t]
\centering\footnotesize
\setlength{\tabcolsep}{4pt}
\begin{tabular}{@{}lrr@{}}\toprule
 & \multicolumn{2}{c}{Training objective} \\
Model & All-token & Target-only \\\midrule
No-memory adapter & 2.3189 & 3.2729 \\
Pooled memory & 2.3190 & 2.4921 \\
Constant input & 2.3106 & 2.4566 \\
\bottomrule\end{tabular}
\caption{Held-out answer-token NLL in nats (lower is better), averaged over ten paired training orders. With target-only training, all three models have higher loss, but the no-memory model worsens more. Thus a relative memory advantage can emerge without lower prediction loss. Appendix~\ref{app:matched-objective-final} retains accuracy, stopping records and uncertainty estimates.}
\label{tab:objective-absolute-main}
\end{table}

With all-token training, pooled memory's advantage over the adapter is uncertain. Under target-only training, it is positive: $+0.7808\;[0.3611,1.2187]$ nats. Constant input also beats the adapter by $+0.8162\;[0.3619,1.2961]$, while the pooled model's advantage over constant input is unresolved: $-0.0354\;[-0.1418,0.0590]$. These crossed, adjusted intervals show an advantage over a baseline that deteriorates more. All target-only losses exceed the 1.8871 NLL of a predictor uniform over eligible candidates, and 43/60 models reach the 1,024-update cap without convergence. Positive utility does not establish reliable state prediction or superiority to constant input (Appendix~\ref{app:matched-objective-final}). Appendix~\ref{app:objective-control} retains the separate earlier experiment.

\section{What a QA Score Gain Establishes}
\label{sec:gold-qa}
A higher score with memory need not mean that the improvement requires the question's document. We test this with trained controls on QASPER and with input changes to the released xRAG model on three QA datasets.

\subsection{Trained Controls on QASPER}
QASPER supplies questions about scientific papers and annotated supporting passages \citep{dasigi2021qasper}. We encode these passages directly, excluding retrieval errors. On 979 test questions, a projector trained on document input scores 19.98 F1, versus 12.54 for the same architecture trained on one constant input. Zeroing either projector leaves the same frozen LLM and prompt, scoring 6.82 (Table~\ref{tab:qasper-zero-main}). Thus zeroing document memory loses 13.16 points, of which 5.73 can also be obtained by a trained input containing no current-document facts. The document-versus-constant advantage remains positive: $+7.43\;[6.15,8.75]$ points. Document content helps, but zeroing damage includes more than that content benefit (Equation~\ref{eq:common-zero}).

\begin{table}[t]
\centering\footnotesize
\setlength{\tabcolsep}{4pt}
\begin{tabular}{@{}lrr@{}}\toprule
Projector training input & Intact F1 & Zeroed F1 \\\midrule
Document & 19.98 & 6.82 \\
Same constant for all documents & 12.54 & 6.82 \\
\bottomrule\end{tabular}
\caption{QASPER F1 on 979 questions, averaged over ten paired training orders. Zeroing the projector output leaves the same frozen LLM and prompt in both rows. Scores fall even when the learned input conveys no current-document facts. Appendix~\ref{app:qasper-details} retains the full results.}
\label{tab:qasper-zero-main}
\end{table}

These models share initialization and ten paired training orders. An expanded learning-rate search retains 92\% of the document-versus-constant development NLL gap (0.334 to 0.308; Appendix~\ref{app:projector-grid}). A separate study trains only decoder LoRA with fixed memory and equal budgets. Training on correct documents gives 21.21 F1, versus 13.73 after training on other documents and 13.27 after training without documents. Both document-trained models receive identical correct memory at evaluation, so this comparison tests training history, not the effect of replacing a document at inference. Appendices~\ref{app:gold-qa}, \ref{app:qasper-details} and~\ref{app:followup-families} retain absolute scores, selection rules and adjusted intervals.

On 405 CoQA first-turn questions \citep{reddy2019coqa}, correct-document training gives 66.56 F1, versus 34.22 for other-document training and 18.86 for training without documents. This short-evidence check supports content benefit; upstream xRAG training includes CoQA, so it does not establish unseen-task transfer (Appendix~\ref{app:coqa-final}).

\subsection{Gains That Do Not Require the Retrieved Passage}
\label{sec:native-generation}
Here we keep the released xRAG-7B weights fixed and compare three inputs: the question's retrieved passage, a passage assigned from another question (a \emph{donor}), and no document. This asks whether a gain over no document is specific to the retrieved passage. Neither retrieval nor donor selection guarantees that a passage contains, or lacks, the answer.

We distinguish three scoring rules. \emph{Match} accepts a normalized answer anywhere in the output. \emph{Exact match} (EM) requires the full normalized output to equal an accepted answer. F1 measures word overlap, penalizing missing and extra words. Extra text can therefore preserve Match while lowering EM or F1. We fixed the software versions, batch size of four, and primary metric, Match, before generating the main outputs; generation details and sensitivity checks are in Appendices~\ref{app:native-xrag} and~\ref{app:native-all48}.

\begingroup
\renewcommand{\footnotesize}{\fontsize{8.5}{9.5}\selectfont}
\renewcommand{\arraystretch}{0.97}
\begin{table}[t]\centering\footnotesize
\setlength{\tabcolsep}{4pt}
\begin{tabular}{lrrr}\toprule
Input & Match (\%) & EM (\%) & F1\\\midrule
Closed book & 57.01 & 1.31 & 19.38 \\
Full text & 65.87 & 1.64 & 21.86 \\
Document & 65.73 & 6.33 & 29.88 \\
Question & 60.53 & 1.66 & 22.43 \\
Zero vector & 57.61 & 1.28 & 19.49 \\
Donor mean & 56.45 & 5.11 & 24.78 \\
\bottomrule\end{tabular}
\caption{Released xRAG on 11,313 TriviaQA questions with fixed software and batch size four. Higher is better. Document encodes the retrieved passage; full text supplies it as text. Closed book omits the vector; question encodes the question instead. Donors average three assignments of other passages (Appendix~\ref{app:native-all48}).}
\label{tab:native-batch-main}
\end{table}

\begin{table*}[t]\centering\footnotesize
\setlength{\tabcolsep}{9pt}
\begin{tabular}{@{}lrrrr@{}}\toprule
 & \multicolumn{3}{c}{Donor minus closed book} & Document minus donor\\
Dataset & Match & EM & F1 & Match\\\midrule
TriviaQA & $-0.57\;[-1.50,0.37]$ & $+3.80\;[3.22,4.42]$ & $+5.40\;[4.64,6.17]$ & $+9.28\;[8.26,10.31]$ \\
NQ-open & $-0.20\;[-1.88,1.44]$ & $+0.14\;[0.04,0.28]$ & $+3.71\;[3.15,4.28]$ & $+8.65\;[6.93,10.42]$ \\
WebQuestions & $+3.99\;[1.49,6.59]$ & $+0.15\;[0.00,0.35]$ & $+3.98\;[2.91,4.84]$ & $+5.22\;[3.05,7.47]$ \\
\bottomrule\end{tabular}
\caption{Score differences with fixed software and batch size four, in percentage or F1 points. Donors are other questions' passages; closed book omits documents. Retrieved passages improve Match over donors in every dataset; donor gains over no document vary. Intervals adjust for fifteen TriviaQA or twelve combined NQ-open/WebQuestions comparisons, not a joint family across datasets.}
\label{tab:native-three}\end{table*}

\endgroup

\paragraph{A gain over no document is not all passage-specific.}
On 11,313 TriviaQA questions \citep{joshi-etal-2017-triviaqa}, we use the repository's ColBERTv2-retrieved passages \citep{santhanam-etal-2022-colbertv2} and three fixed donor assignments. F1 is 19.38 without a document, 24.78 with donors, and 29.88 with the retrieved passage (Table~\ref{tab:native-batch-main}). The two improvements answer different questions: donors beat no document by 5.40 points, and the retrieved passage beats donors by a further 5.10. The first improvement does not require the question's retrieved passage; the second supports an advantage for that passage over the assigned alternatives. Neither contrast identifies a causal fraction of information recovered.

The scoring rule changes the conclusion about donors. Their EM gain is $+3.80\;[3.22,4.42]$ points, but their Match difference is uncertain: $-0.57\;[-1.50,0.37]$. The retrieved passage beats donors by $+9.28\;[8.26,10.31]$ Match points. These intervals adjust across fifteen comparisons. Thus higher word overlap or exact match need not establish that more answers contain an accepted answer.

\paragraph{The pattern extends, but not every metric agrees.}
We repeat the fixed-model comparison on 3,610 NQ-open and 2,032 WebQuestions examples, using their cached DPR passages \citep{kwiatkowski-etal-2019-natural,lee-etal-2019-latent,berant-etal-2013-semantic,karpukhin-etal-2020-dense}. Donor F1 gains over no document are positive on all three datasets, and the retrieved passage beats donors on Match throughout (Table~\ref{tab:native-three}). Donor Match is uncertain on NQ-open but positive on WebQuestions; donor EM gains on these two datasets are near zero. The additional datasets use a separate twelve-comparison adjustment (Appendix~\ref{app:native-two-datasets}).

\paragraph{Training changes sensitivity within QASPER.}
An earlier fixed-memory study permits a direct check with questions, memory bytes, donor assignments, prompt and generation rule held fixed. Before decoder training, correct and other-paper memory score 14.21 and 10.90 F1; after document training, they score 21.29 and 7.15. Their gap grows from 3.31 to 14.14 points, an increase of $10.83\;[9.07,12.59]$ under the original simultaneous interval over training orders and papers. Training improves answers with correct memory while reducing performance with other-paper memory. This is evidence that training changes the response to document content within the same dataset.

A post hoc comparison of saved public-model QASPER outputs also finds 10.90 F1 with other-paper memory versus 6.60 without a document. Thus the donor-gain pattern is not absent from QASPER before adaptation. These outputs come from separate runs with matching recorded checkpoint and generation code; removing the document also changes the prompt, and complete runtime equivalence is not established (Appendix~\ref{app:qa-reconciliation}).

QASPER supplies annotated evidence, whereas the released-model study uses retrieved passages; their answer-generation procedures also differ. These differences, together with scoring, can affect measured gains, but our studies do not isolate each one's contribution. The supported conclusion is that training and evaluation choices change what a score difference establishes, not that one QA dataset benefits from memory and another does not.

\section{Supporting Lexical Checks}
\label{sec:xrag-content}
To test content benefit beyond QA, we predict words that occurred earlier in the text. Three xRAG projectors receive the earlier text, the recent suffix, or a constant, while SFR and the decoder remain frozen. Every decoder sees the same 281-token suffix. The constant is the mean of 8,192 training-prefix embeddings. We select each input's learning rate and checkpoint on 512 development targets; all choose the highest tested rate, $2\times10^{-4}$. Ten paired training orders use 512 updates at batch size 64 (Appendix~\ref{app:xrag-content}).

\paragraph{Positive lexical benefit.}
On 3,044 PG-19 repeated-word targets, document input gives 4.5088 NLL versus 4.8145 for constant and 4.8254 for suffix input. The document advantage is $+0.3166$ nats over suffix input and $C=+0.3057\;[0.2530,0.3585]$ over constant input. Document input also improves accuracy over constant input by $+2.13\;[1.15,3.09]$ points. The crossed simultaneous lower bound is 0.85 points, below the one-point descriptive benchmark. Zeroing constant input raises NLL by $+1.2360$ nats, although its advantage over suffix input is uncertain: $+0.0109\;[-0.0042,0.0259]$. This again separates reliance on a learned input from an advantage over another trained input.

\paragraph{A harder case.}
We construct 217 pairs from 67 NASA articles by replacing a repeated word with one of equal token length while preserving the local suffix. On original inputs, document input improves NLL over constant input by $C=+0.2098$ nats, but the crossed simultaneous interval $[-0.0058,0.4424]$ includes zero. The adjusted accuracy difference is $-2.86\;[-8.85,2.70]$ points. After substitution, document-input accuracy falls from 47.56\% to 0.23\%, despite a positive NLL contrast. The model does not reliably recover the substituted words; these probes do not establish factual recovery (Appendix~\ref{app:xrag-content}).

\section{Discussion}
\label{sec:discussion}
Three results expose a common inference problem. Relative gains can grow while both models worsen (Table~\ref{tab:objective-absolute-main}). Removing a learned constant input can hurt despite that input conveying no current-document facts (Table~\ref{tab:qasper-zero-main}). Other questions' passages can raise QA scores (Table~\ref{tab:native-three}). A score change alone therefore cannot identify the benefit of the current document's content.

Evaluations should match claims to comparisons. Zeroing tests reliance after training; a separately trained no-memory model tests performance benefit. Trained constant-input controls and replacement passages test whether gains require document-dependent input or the question's passage. Our positive QA results support document benefit while showing why that benefit cannot be inferred from every observed gain.

\section{Conclusion}
Zeroing memory after training can hurt even when a model trained without it performs better. Likewise, a relative advantage can grow while both models' prediction losses worsen. Our QA comparisons show that document content helps, yet some gains over zeroed memory or no document also occur without the question's document. Within QASPER, decoder training increases sensitivity to correct versus other-document memory; in released-model QA, generation and scoring choices change the conclusions about other-passage gains. Before attributing gains to document content, evaluations should name the comparison model, report absolute performance, and specify how answers are generated and scored.

\FloatBarrier
\clearpage
\section*{Limitations}
Every benefit in this paper is relative to a specified comparison model, task, and training procedure. We do not identify the best performance an architecture could achieve, establish convergence, or match total pretraining and inference costs. Appendix~\ref{app:external-du} and the evidence archive retain selection rules and candidate settings. The comparison with shared initialization and equal development-search opportunities does not establish that zeroing hurts while the trained alternative performs better. The longer training study also does not compare independently optimal settings. Optimization differences may matter, but these results do not isolate them as a causal explanation.

The generated stories permit copying of repeated values. Predicting one distinguishing token with preceding gold text supplied is easier to constrain than generating an entire state value. PG-19 repeated words test lexical prediction rather than annotated facts. The restricted-attention experiments deliberately hide earlier text from direct attention and do not represent every deployment setting. QASPER supplies annotated evidence, so its results do not evaluate retrieval or show that those passages are uniquely necessary. Its 128-question development panel and ten training orders provide limited optimization coverage. Masking evidence and matched non-evidence text controls length and approximate position, but not semantics or all changes in the input distribution.

The studies use different tokenizers, memory modules, and multiple-comparison adjustments. We compare paired conditions within each study and do not rank raw NLL across tokenizers. Separating development and test data also does not exclude foundation-model pretraining exposure. The xRAG decoder or SFR encoder may have encountered PG-19, and an embedding could help identify familiar text as well as convey current context. Pythia was trained on the Pile, which includes PG-19 \citep{biderman2023pythia,gao2020pile}; our local holdout does not remove that exposure. Selecting settings on folds separate from those used for scoring provides internal validation only.

Intervals that resample items while holding trained models fixed do not cover training uncertainty. Crossed intervals over ten training orders and documents include order variation at a shared initialization, but not variation across arbitrary initializations or the full configuration search. Finite bootstrap samples also limit precision in the tails used for simultaneous intervals. An interval containing zero is not proof of no effect.

The NASA task uses automatic word substitutions without human validation of semantic answers. Holding local suffixes fixed restricts the change to earlier input, but the large accuracy difference shows that the original and substituted examples have different difficulty. More generally, constant input restricts what the projector can compute even when its architecture is unchanged. The document-versus-constant difference therefore describes the trained models, not a unique causal fraction of utility.

The released-model TriviaQA analysis uses one frozen checkpoint and conditions on the supplied retrieval and three fixed donor assignments. Encoding the question adds computation to a prompt that already contains it; this changes evaluation input without retraining. Although fixed software and batch size four closely reproduce reported aggregate Match, the original per-question outputs and complete configuration are unavailable. The question-input gain and its ratio to the document-input gain are post hoc descriptions outside the corrected 15-effect family. Changes in scores across batches do not isolate output length as their sole cause.

\section*{Ethics Statement}
We use synthetic templates and public texts and datasets, including TriviaQA, NQ-open, WebQuestions, PG-19, QASPER, CoQA, MuSiQue, and NASA articles. The supporting rank study uses LongBench \citep{bai2023longbench}. No participants were recruited or new human annotation commissioned. Data and model use follows upstream terms, and public access does not imply unrestricted redistribution. The numerical supplement preserves complete comparison families and experiment settings, including results that qualify the claims. It separates author-generated evidence from third-party checkpoints and corpora.

\subsection*{AI Assistance}
Generative AI assisted literature search, study-design discussion, implementation and debugging, experiment orchestration, analysis, manuscript writing, and figure preparation. The authors checked input and checkpoint identities, development selections, saved-answer rescoring, paired effect reconstruction, and the cited source passages. They reviewed the rendered PDFs and remain responsible for the research and its presentation.

For the literature audit, AI tools assisted full-text review, initial classification, a second coding pass, and adjudication of disagreements. The reported agreement statistics describe these AI-assisted coding passes, not agreement between independent human annotators.

\bibliography{custom}
\appendix
\clearpage
\raggedbottom
\section{Controlled Probe: Implementation and Data}
\label{app:method-details}\label{app:controlled-source}\label{sec:plangen}
\paragraph{Local access and delayed memory.}
This probe asks whether a model can use compressed records of earlier story events when those events are outside its local attention window. At each layer, token $t$ attends only to itself and the preceding $k-1$ tokens. With $L$ layers, information can therefore travel at most $L(k-1)$ positions through local attention. We divide the sequence into blocks of $r=16$ consecutive tokens. Token $x_t$ belongs to block $b(t)=\lfloor(t-1)/r\rfloor$, where the floor rounds down. For each completed block, we average its hidden states to obtain $\bar h_j$. The encoder $f_\phi$ applies layer normalization, an $H\to H$ linear map, GELU and an $H\to256$ linear map, where $H$ is the backbone's hidden width. The resulting memory slots satisfy $\hat p_i=f_\phi(\bar h_{i-2})$ for $i\ge2$. This two-block delay prevents a token from reading a summary of its own unfinished block. Slot 0 is learned, slot 1 is $f_\phi(0)$, and the memory bank resets within each evaluation window.

Pooled reading averages all available projected slots and adds their gated contribution to the local hidden state:
\begin{align}
z_t&=\frac{1}{b(t)+1}\sum_{i\le b(t)}(W_p\hat p_i+b_p),\\
g_t&=\sigma(W_g[h_t;z_t]+b_g),\qquad
\tilde h_t=h_t+g_t\odot z_t.
\end{align}
Here $h_t$ is the local hidden state, $\sigma$ is the sigmoid function, $[\,;\,]$ concatenates vectors, and $\odot$ multiplies them elementwise. $W_p,b_p$ project each slot to the hidden width, and $W_g,b_g$ determine the gate. The output head predicts next-token probabilities $Q_\theta(x_{t+1}\mid x_{1:t},\hat P_{\le b(t)};\mathcal M_k)$, where $\mathcal M_k$ is the local attention mask and $\hat P_{\le b(t)}$ contains the available memory slots. The indexed variant reads only the current slot, using $z_t=W_p\hat p_{b(t)}+b_p$; its conservative reach adds $3r=48$ positions. Scored targets exclude recurrences within 384 tokens, placing the relevant earlier occurrence beyond both local and indexed reach. We check that changing future tokens leaves earlier predictions unchanged. Block means and slots remain attached to the computation graph, so LoRA receives gradients through both local and memory paths.

\paragraph{Comparison models and training.}
The no-memory comparison model adds a residual multilayer perceptron (MLP) at the current position, with layer normalization, a linear map, GELU and a second linear map. This local adapter cannot access the memory bank. Its expansion width matches the memory module's parameter count within 0.06\% (Table~\ref{tab:state-settings}); this does not equalize the functions the modules can represent or their inference costs. We separately train four conditions: local LoRA alone, LoRA with the local adapter, pooled memory and indexed memory. They share the backbone, tokenizer, data, objective and LoRA configuration. The 72 fits use seeds 71--73, learning rate $10^{-4}$, cosine decay and 183 updates of 262,144 tokens (47,972,352 total). Validation runs every 8M tokens, but evaluation uses the fixed final checkpoints. LoRA rank and scale are 16/32. LoRA adapts fused QKV/MLP projections for Pythia, and q/k/v plus gate/up/down projections for Llama. The base model weights remain frozen.

\begin{table}[!t]
\centering\small
\setlength{\tabcolsep}{3pt}
\begin{tabular}{@{}lrrrrr@{}}
\toprule
Backbone & $L$ & $k$ & $H$ & MLP width & Module (M) \\
\midrule
Pythia-2.8B & 32 & 10 & 2,560 & 4,098 & 20.98 \\
Pythia-6.9B & 32 & 10 & 4,096 & 6,402 & 52.45 \\
Pythia-12B & 36 & 8 & 5,120 & 7,938 & 81.30 \\
Llama-3.2-1B & 16 & 20 & 2,048 & 3,330 & 13.64 \\
Llama-3.2-3B & 28 & 11 & 3,072 & 4,866 & 29.90 \\
Llama-3.1-8B & 32 & 10 & 4,096 & 6,402 & 52.45 \\
\bottomrule
\end{tabular}
\caption{Controlled-state settings: layer count $L$, window $k$, backbone width $H$, and adapter intermediate width from saved tensors. Module counts exclude common LoRA. Parameter-count matching error is below 0.06\%.}
\label{tab:state-settings}
\end{table}

\begin{table*}[!t]
\centering\small
\setlength{\tabcolsep}{4pt}
\begin{tabular}{@{}lrrrcc@{}}
\toprule
Backbone & Adapter NLL & Pooled NLL & Zeroed NLL & Gain $U$ [95\% CI] & Zeroing $D$ [95\% CI] \\
\midrule
Pythia-2.8B & 2.4284 & 2.5067 & 3.0140 & $-0.0784\;[-0.0899,-0.0665]$ & $+0.5073\;[+0.4814,+0.5337]$ \\
Pythia-6.9B & 2.3499 & 2.4293 & 2.7966 & $-0.0795\;[-0.0923,-0.0671]$ & $+0.3673\;[+0.3441,+0.3909]$ \\
Pythia-12B & 2.3736 & 2.4459 & 2.7628 & $-0.0723\;[-0.0849,-0.0595]$ & $+0.3169\;[+0.2946,+0.3394]$ \\
Llama-3.2-1B & 2.3019 & 2.3801 & 2.8610 & $-0.0782\;[-0.0909,-0.0655]$ & $+0.4809\;[+0.4539,+0.5073]$ \\
Llama-3.2-3B & 2.1926 & 2.3343 & 2.7986 & $-0.1417\;[-0.1574,-0.1265]$ & $+0.4643\;[+0.4395,+0.4892]$ \\
Llama-3.1-8B & 2.0586 & 2.0910 & 2.1906 & $-0.0325\;[-0.0400,-0.0248]$ & $+0.0996\;[+0.0848,+0.1145]$ \\

\bottomrule
\end{tabular}
\caption{Pooled PlanGen on state tokens (nats; marginal 95\% document-bootstrap CIs, three fitted seeds). $U$ compares pooled memory with the parameter-matched adapter; $D$ compares its intact and zeroed inputs. All six have $U<0,D>0$. Candidate accuracy does not establish state recovery; mean uniform-candidate references are 11.35\% (Pythia) and 17.24\% (Llama). Table~\ref{tab:controlled-all-arms} reports all arms.}
\label{tab:harder-scale}
\end{table*}

\paragraph{Stories and scoring.}
The generator creates 6,400/800/800 train/dev/test stories by assigning values to entity attributes and mentioning them later. Each story has six assignment-to-mention sequences, including two updates to earlier assignments. However, the recurrence filter removes all test targets governed by those updates. Training and development stories are joined with end-of-sequence (EOS) tokens in groups of 64; test windows stay within individual documents. At evaluation, the model receives the preceding gold text. We score the first token that distinguishes the true value from its allowed alternatives, requiring at least two candidate tokens. The earlier mention must lie within 2,000 tokens, and the scored token must be absent from the preceding 384. These rules retain 3,847 Pythia and 3,687 Llama targets. Every model within a backbone is scored on the same targets.

Negative log-likelihood (NLL) uses probabilities over the full vocabulary. In contrast, the accuracy in Table~\ref{tab:controlled-all-arms} chooses among the permitted candidate tokens. A predictor that chooses uniformly among an item's $K$ candidates has mean accuracy 11.35\% for Pythia and 17.24\% for Llama; its mean item-wise $\log K$ is 2.233918 and 1.902385 nats, respectively. These benchmarks already know the candidate set. They are neither uniform predictions over the full vocabulary nor model NLLs renormalized over candidates. The models' low absolute accuracy does not establish reliable recovery of the story state.

\begin{table*}[!t]
\centering\small
\setlength{\tabcolsep}{5pt}
\begin{tabular}{@{}lrrrr@{}}
\toprule
Backbone & LoRA only & Local adapter & Pooled & Indexed \\
\midrule
Pythia-2.8B & 2.6062 / 11.26 & 2.4284 / 10.86 & 2.5067 / 11.32 & 2.5168 / 11.29 \\
Pythia-6.9B & 2.4753 / 11.28 & 2.3499 / 11.00 & 2.4293 / 10.99 & 2.4261 / 11.99 \\
Pythia-12B & 2.5359 / 11.62 & 2.3736 / 11.44 & 2.4459 / 11.00 & 2.4516 / 11.11 \\
Llama-3.2-1B & 2.6037 / 13.54 & 2.3019 / 13.89 & 2.3801 / 13.77 & 2.4015 / 13.79 \\
Llama-3.2-3B & 2.6051 / 13.90 & 2.1926 / 14.48 & 2.3343 / 14.07 & 2.3449 / 14.11 \\
Llama-3.1-8B & 2.1252 / 13.71 & 2.0586 / 13.62 & 2.0910 / 13.63 & 2.0977 / 13.68 \\

\bottomrule
\end{tabular}
\caption{Controlled state targets: full-vocabulary NLL / candidate-token accuracy (\%), averaged over seeds 71--73 on the same positions as Table~\ref{tab:harder-scale}. All four arms are shown; raw NLLs are compared only within a backbone.}
\label{tab:controlled-all-arms}
\end{table*}

\paragraph{Incorrect candidate values in local text.}
We use saved Llama-3B outputs to ask whether low accuracy is confined to targets with a misleading candidate nearby. An incorrect candidate occurs in the 281-token local input for 860 of 3,687 targets. Pooled-memory accuracy is 13.18\% with such a candidate and 14.34\% without one; corresponding uniform-candidate benchmarks are 14.55\%/18.07\%. Adapter accuracy is 12.98\%/14.94\%. Low accuracy therefore persists even without a local distractor, although this association does not identify the cause. Because the original files omit total probability assigned to candidates, a post hoc forward audit records it for 512 fixed targets. Both full-vocabulary and candidate-restricted pooled-memory accuracy are 14.78\%, compared with 18.55\% for a predictor based only on training frequencies. The crossed interval for the accuracy difference is $[-7.94,0.39]$ points. The recurrence filter excludes all 92 targets for the generated value \texttt{route-seven}, because the scored token occurs locally. Supplementary Appendix~S27 gives the complete diagnostic.

\paragraph{Interventions and diagnostic scope.}
Zeroing sets memory values to zero but retains projection bias; bypassing instead removes the added residual contribution. A bank swap supplies memory from a different document with the same number of visible slots. Its intact and swapped scores use exactly the same covered subset. Permuting slots cannot change their pooled mean. Zeroing only document-summary slots nearly reproduces the damage from zeroing all slots. In contrast, zeroing only initial slots, replacing document summaries with their training mean, and swapping banks change NLL by amounts within $\pm0.01$ nat. Supplementary Table~S3 reports all seven contrasts and their subset sizes. It also preserves the small absolute-NLL differences between H100 runs at batch eight and A6000 runs at batch one. Each contrast pairs BF16 measurements within the same execution. These measurements describe which inference inputs matter to the trained model; neither rounding differences nor slot interventions identify how training produced that dependence. Supplementary Section~S1 gives the architecture diagram and full intervention settings.

\section{Earlier Objective and Constant-Input Controls}
\label{app:objective-control}
This study asks whether an apparent memory advantage changes when training emphasizes the scored state tokens. It uses three seeds and a fixed budget of 512 updates. Appendix~\ref{app:matched-objective-final} reports a separate ten-seed study with equal development searches for each model and objective. The sign reversal observed here must not be transferred to that later recipe.
\label{app:content-control}\label{app:constant-interpretation}
\paragraph{Matched training and selection.}
We train pooled memory and a local adapter under two objectives: predicting every next token, or predicting only the designated state tokens. Both objectives receive identical input sequences, and the all-token objective includes the state targets. All conditions use public Llama-3.2-3B, attention window 11 and rank-16/scale-32 LoRA. Pooled memory and the adapter contain 29,900,288/29,898,496 module parameters. The adapter width here is 4,864; the separately matched configuration in Appendix~\ref{sec:plangen} uses 4,866. These are distinct saved tensor widths. Initialization and input order are paired within each seed, and the residual gate starts nonzero.

Eight development runs at seed 301 compare learning rates $5\times10^{-5}$ and $2\times10^{-4}$ for 128 updates. On a fixed panel of 128 development items, the mean endpoint NLL across the four model/objective conditions is 4.2523/3.4885. This selects the common rate $2\times10^{-4}$. Final fits at seeds 311--313 run 512 updates with batch four, zero weight decay, AdamW $\beta=(0.9,0.95)$, gradient clipping at one, 16 warmup updates and cosine decay to one tenth of the peak rate. They use FP32 trainable parameters and BF16 base weights on A6000. Evaluation uses the fixed final checkpoint. Each fit trained only on state targets receives 2,048 sampled presentations. Development curves neither select test checkpoints nor establish convergence.

\paragraph{Constant input throughout training.}
To test whether document content accounts for the observed advantage, six additional pooled-memory fits receive the same vector in place of every document block. We compute that vector once as the mean of 190,938 complete delayed training blocks, using the initial masked backbone, FP64 accumulation and FP32 storage. These fits match the normal-memory fits in initial tensors, sampled examples, gate, slots, delay and budget. The encoder and gate remain trainable, and local hidden states and sequence length remain available. Thus the memory path is independent of document content, but the model can still make different predictions at different positions. Replacing document values also removes the gradient path through remote document states. The comparison therefore changes more than the information in an otherwise identical optimization problem.

The shared objective panel has 1,024 targets in 617 stories. The content comparison evaluates all eighteen fits on 1,024 targets from 588 of 800 new stories (generator seed 2026101041), with no exact full-story/window overlap with the earlier pools. It uses the same generator and causal selection rules. Nine unmatched items are excluded only from bank-swap contrasts. Table~\ref{tab:content-control} gives every new-document effect, including accuracy and direct objective differences; Supplementary Section~S2 retains the shared-panel results and development records.

\begin{table*}[!t]
\centering\small
\setlength{\tabcolsep}{5pt}
\begin{tabular}{@{}llrr@{}}
\toprule
Training loss & Model & Target NLL & Accuracy (\%) \\
\midrule
All-token & Local adapter & 2.1775 & 12.92 \\
All-token & Normal pooled memory & 2.2068 & 13.64 \\
All-token & Document-independent memory & 2.2055 & 13.48 \\
Target-only & Local adapter & 2.2422 & 16.83 \\
Target-only & Normal pooled memory & 2.1907 & 16.47 \\
Target-only & Document-independent memory & 2.1869 & 16.70 \\
\bottomrule
\end{tabular}
\medskip
\setlength{\tabcolsep}{3pt}
\begin{tabular}{@{}lccc@{}}
\toprule
Effect & All-token & Target-only & Target minus all-token \\
\midrule
$U$ (nats) & $-0.0293\;[-0.0443,-0.0146]$ & $+0.0515\;[+0.0350,+0.0681]$ & $+0.0808\;[+0.0583,+0.1033]$ \\
$D_{\rm zero}$ (nats) & $+0.1067\;[+0.0789,+0.1348]$ & $+0.0617\;[+0.0438,+0.0796]$ & $-0.0450\;[-0.0765,-0.0138]$ \\
$D_{\rm swap}$ (nats) & $+0.00018\;[-0.00073,+0.00108]$ & $-0.00008\;[-0.00018,+0.00003]$ & $-0.00025\;[-0.00116,+0.00065]$ \\
$U_{\rm acc}$ (pp) & $+0.72\;[-0.45,+1.88]$ & $-0.36\;[-1.46,+0.74]$ & $-1.07\;[-2.73,+0.58]$ \\
$U_{\rm const}$ (nats) & $-0.0280\;[-0.0429,-0.0131]$ & $+0.0553\;[+0.0399,+0.0709]$ & $+0.0834\;[+0.0622,+0.1047]$ \\
$D_{\rm const}$ (nats) & $+0.1110\;[+0.0847,+0.1376]$ & $+0.1047\;[+0.0809,+0.1284]$ & $-0.0063\;[-0.0408,+0.0284]$ \\
$U_{\rm const,acc}$ (pp) & $+0.55\;[-0.59,+1.70]$ & $-0.13\;[-1.19,+0.93]$ & $-0.68\;[-2.34,+0.96]$ \\
$C$ (nats) & $-0.0013\;[-0.0055,+0.0029]$ & $-0.0038\;[-0.0144,+0.0067]$ & $-0.0025\;[-0.0141,+0.0089]$ \\
$C_{\rm acc}$ (pp) & $+0.16\;[-0.46,+0.80]$ & $-0.23\;[-1.17,+0.71]$ & $-0.39\;[-1.54,+0.75]$ \\
\bottomrule
\end{tabular}
\caption{New-document content control: 1,024 targets from 588 stories, A6000, seeds 311--313. All 27 conditional 95\% intervals are shown. $U$ compares normal memory with the local adapter; $U_{\rm const}$ compares document-independent memory with the same local adapter. $D_{\rm const}$ is zeroing damage to that control. $C=\mathrm{NLL}_{\rm const}-\mathrm{NLL}_{\rm normal}$ measures normal memory\textquotesingle s advantage; $C_{\rm acc}$ uses normal-minus-constant accuracy. Swap covers 1,015 targets; the other effects cover all 1,024. Accuracy is full-vocabulary first-token accuracy, unlike the candidate-restricted accuracy in Table~\ref{tab:controlled-all-arms}; its contrasts use percentage points. An uninformed uniform full-vocabulary predictor has accuracy $100/|V|$ percent, not the candidate-aware $100/K$ reference.}
\label{tab:content-control}
\end{table*}

Both document memory and the constant-input path change from worse than the local adapter to better than it when training switches to designated targets. Zeroing still harms both trained paths. Under all-token training, the NLL difference between document and constant input lies within $\pm0.01$ nat; under target-only training, its interval includes zero. The accuracy contrasts also include zero. The 0.01-nat benchmark is about 34\%/19\% of the absolute document-memory utility estimates, so lying within this benchmark does not establish that content contributes negligibly. Both the memory model and its adapter comparator change when the objective changes. Their relative gain cannot assign a causal contribution to either change. The supporting evidence retains all 27 effects and their Bonferroni sensitivity bounds.

\section{Gold-Answer QA: Data, Fitting and Contrasts}
\label{app:gold-qa}\label{app:em-followup}
\paragraph{Eligible inputs and independent paper splits.}
This experiment tests compression when the relevant evidence is supplied, without introducing retrieval errors. QASPER v0.3 provides 1,458/667/979 eligible train/dev/test questions from 721/260/388 disjoint papers. We retain answerable, non-boolean annotations whose evidence strings match source paragraphs after whitespace normalization, then join all matched evidence paragraphs in document order. Evidence spans contain 16--2,048 encoder tokens; questions contain at most 128 decoder tokens; the canonical answer plus EOS contains at most 64. Extractive answer spans are joined with commas; other cases use the annotated free-form answer. All accepted answer aliases remain available for scoring, while the first is the training target. A fixed hash of question IDs orders the items and selects the first 128 development questions for model selection. Because inputs contain annotated evidence, this experiment evaluates neither retrieval nor reading an entire paper.

\paragraph{Three trained inputs.}
We train three projectors to compare evidence content with other inputs through the same interface. All start from the public 33,562,624-parameter xRAG projector; the SFR encoder and Mistral decoder remain frozen. The projector receives an evidence embedding, a question embedding, or a constant equal to the mean training-evidence embedding. The constant uses FP64 accumulation, FP32 rounding and BF16 projector input. In every condition, the decoder receives one projected vector plus the same question text. Although all projectors have the same nominal parameter count, the constant condition can produce only one learned 4,096-dimensional vector. It therefore has less input-dependent capacity.

Each input condition independently screens learning rates $10^{-5},5\times10^{-5},2\times10^{-4}$ at seed 491, using batch 16, zero weight decay and 128 updates. Minimum development NLL across checkpoints 0/32/64/128 selects the rate; ties favor the lower rate and then the earlier checkpoint. All conditions select the grid maximum, $2\times10^{-4}$. Final training uses ten seeds 501--510 with identical initial tensors and paired example permutations. Each fit runs 512 updates and selects the checkpoint with minimum development NLL among 0/128/256/384/512. AdamW uses $\beta=(0.9,0.95)$, gradient clipping at one, eight warmup updates and cosine decay to one tenth of the peak rate. Training averages canonical-answer token losses, including EOS, within each question and then across questions. This finite search does not establish an optimum beyond the tested grid.

\paragraph{Generation and controls.}
The prompt contains \emph{Background: \texttt{<xRAG>}}, the question, and \emph{The answer is:}. The full-text comparison replaces the vector marker with evidence text; the closed-book comparison omits background and receives only the question. Greedy generation stops at EOS or 96 new tokens, rather than at punctuation. We compute normalized token-overlap F1 and exact match (EM), taking the best score over accepted answer aliases. NLL uses the canonical answer wording. Before test evaluation, a development check requires full-text F1 of at least 20 and a lead of five points over closed-book F1. Measured full-text, closed-book and public compressed-model F1 is 25.59/7.13/14.74. This check establishes that the supplied evidence is usable by the full-text model; it does not require the compressed model to outperform either comparison.

Zeroing sets the projected vector to zero, while swapping replaces it with a fixed embedding from another paper. After zeroing, the document-trained and constant-trained projector conditions give identical predictions on all 9,790 seed--question pairs. Their shared zero-input score therefore cancels when comparing the two zeroing effects, both itemwise and after paired averaging; the separately estimated interval endpoints need not add. Table~\ref{tab:qa-content-effects} reports all twelve effects, and Table~\ref{tab:qa-content-absolute} reports absolute scores. The artifact preserves every seed, development trajectory and selection. Secondary EM gains from the same answers are $U_{\rm question}=+1.14\;[0.40,1.91]$ for document versus question input and $C=+1.16\;[0.46,1.91]$ for document versus constant input. Zeroing and swapping lower document-input EM by $+4.29\;[3.20,5.43]$ and $+2.85\;[1.89,3.86]$ points. In the same order, bounds that also resample training-order seeds and adjust across this separate four-effect family are $[0.19,2.10]$, $[0.24,2.09]$, $[2.91,5.67]$ and $[1.62,4.08]$.

\begin{table*}[t]\centering\small
\begin{tabular}{@{}lrr@{}}\toprule
QA effect (estimate) & Fixed-fit percentile 95\% & Crossed simultaneous 95\% \\\midrule
$U$ (F1) (+7.4436) & $[+6.1977,+8.7125]$ & $[+5.4742,+9.4130]$ \\
$C$ (F1) (+7.4336) & $[+6.1472,+8.7450]$ & $[+5.4782,+9.3889]$ \\
$D_{\rm zero}$ (F1) (+13.1594) & $[+11.6620,+14.6691]$ & $[+10.9762,+15.3426]$ \\
$D_{\rm swap}$ (F1) (+12.8842) & $[+11.5753,+14.2269]$ & $[+10.8939,+14.8745]$ \\
Constant zeroing (F1) (+5.7258) & $[+4.3961,+7.0852]$ & $[+3.7904,+7.6612]$ \\
Constant minus question (F1) (+0.0100) & $[-0.4945,+0.5263]$ & $[-0.8445,+0.8645]$ \\
$U$ (NLL) (+0.3357) & $[+0.3033,+0.3693]$ & $[+0.2885,+0.3830]$ \\
$C$ (NLL) (+0.3182) & $[+0.2858,+0.3515]$ & $[+0.2713,+0.3650]$ \\
$D_{\rm zero}$ (NLL) (+8.5873) & $[+8.2558,+8.9295]$ & $[+8.1115,+9.0631]$ \\
$D_{\rm swap}$ (NLL) (+0.9528) & $[+0.9043,+1.0028]$ & $[+0.8643,+1.0412]$ \\
Constant zeroing (NLL) (+8.2692) & $[+7.9405,+8.6088]$ & $[+7.7975,+8.7408]$ \\
Constant minus question (NLL) (+0.0176) & $[+0.0122,+0.0231]$ & $[+0.0081,+0.0271]$ \\
\bottomrule\end{tabular}
\caption{All twelve trained-projector QA effects. Conditional percentile intervals hold ten fits fixed; crossed sensitivity resamples paired seeds and papers with adjustment across these twelve effects. Positive $U/C$ favor document input, and positive intervention scores mean damage. F1 uses points and NLL nats. Fixed-storage LLM bounds appear in Tables~\ref{tab:reader-main} and~\ref{tab:fixed-reader-effects}; lexical bounds in Table~\ref{tab:xrag-content-fresh-nasa}.}
\label{tab:qa-content-effects}\label{tab:qa-core-sensitivity}\end{table*}

\section{LLM Adaptation at Fixed Storage}
\label{supp:fixed-reader}\label{app:direct-contrasts}
\paragraph{Fixed document representation and matched adaptation.}
This experiment asks whether training the answer-generating LLM can improve answers while leaving compressed storage unchanged. The public SFR encoder and projector save one 4,096-dimensional BF16 vector (8,192 bytes) per evidence input. We reuse these exact bytes without training the encoder or projector, re-encoding evidence, or passing evidence text to the decoder. One decoder trains on the correct document vector; the other trains on the public projector's output for the mean training embedding. This second condition trains the decoder with constant input. It is distinct from Appendix~\ref{app:gold-qa}, which trains a projector while keeping the decoder frozen.

Only decoder LoRA parameters are trained. They use rank 16, alpha 32 and zero dropout on attention q/k/v/o and MLP gate/up/down projections, totaling 41,943,040 parameters. Both conditions use the same QASPER inputs and paired seeds 501--510. Their LoRA initialization has zero B matrices and identical A matrices. They share 512 updates, batch 16, zero weight decay and greedy answers capped at 96 tokens. Separate development screens run for 128 updates at seed 491 over the QA learning-rate grid; both select $5\times10^{-5}$. Final development NLL is evaluated at 0/128/256/384/512, and every fit selects step 128. Decoder adaptation adds trainable parameters and training cost while preserving storage. The mean-trained decoder receives one fixed vector throughout training but varying document vectors in the intact evaluation. Its comparison with the document-trained decoder therefore includes this change in input distribution.

\paragraph{Paired comparisons.}
We compare improvement over the public decoder, improvement over a decoder trained with constant input, and the extra improvement obtained with the correct document rather than a donor. For public, document-trained and mean-trained LLMs $b,d,a$, and intact, swapped and mean inputs $i,s,m$, define $R=d_i-b_i$, $T_{\rm LLM}=d_i-a_m$ and $I_{\rm LLM}=(d_i-b_i)-(d_s-b_s)$. The study's joint claim requires all three fixed-fit 95\% lower bounds to exceed zero. The additional effects are $J=(d_i-d_s)-(a_i-a_s)$, $D_{\rm swap}=d_i-d_s$, $D_{\rm mean}=d_i-d_m$, and $A=a_m-b_m$. $J$ compares swap sensitivity between the two training conditions; $A$ measures improvement from training on constant input. A donor is the next item from a different paper in a fixed ordering, chosen independently of outcomes. The tables report all nine conditions and seven effects. The 63 evaluations retain 61,677 answers.

\begin{table*}[t]
\centering\small
\begin{tabular}{lrcc}
\toprule
F1 contrast & Estimate & Fixed-fit paper 95\% & Crossed simultaneous 95\% \\
\midrule
$R$: intact recovery & $+7.08$ & $[+5.74,+8.44]$ & $[+5.22,+8.95]$ \\
$T_{\rm LLM}$: trained-reference gain & $+8.87$ & $[+7.62,+10.12]$ & $[+7.11,+10.63]$ \\
$I_{\rm LLM}$: intact-minus-donor recovery & $+10.83$ & $[+9.55,+12.11]$ & $[+9.07,+12.59]$ \\
$J$: training specificity & $+9.95$ & $[+8.84,+11.05]$ & $[+8.37,+11.52]$ \\
$D_{\rm swap}$: swap damage & $+14.14$ & $[+12.81,+15.46]$ & $[+12.32,+15.95]$ \\
$D_{\rm mean}$: mean damage & $+10.99$ & $[+9.76,+12.20]$ & $[+9.24,+12.73]$ \\
$A$: document-free adaptation & $+0.19$ & $[-0.94,+1.33]$ & $[-1.44,+1.82]$ \\
\bottomrule
\end{tabular}
\caption{All seven fixed-memory LLM effects, in F1 points. Each focal contrast in this earlier study ($R,T_{\rm LLM},I_{\rm LLM}$) is positive under both constructions. The seven-effect simultaneous family also resamples training-order seeds; neither interval covers arbitrary initialization or recipe selection. Two F1 points is a descriptive magnitude benchmark, not the existence criterion.}
\label{tab:fixed-reader-effects}
\end{table*}
\begin{table*}[t]
\centering\small
\begin{tabular}{lrrrr}
\toprule
LLM and memory & F1 & EM (\%) & NLL & Tokens \\
\midrule
Public LLM, intact & 14.21 & 1.12 & 3.4513 & 48.90 \\
Public LLM, swap & 10.90 & 0.92 & 3.9922 & 48.47 \\
Public LLM, mean & 12.23 & 0.82 & 3.6518 & 40.51 \\
Document-trained LLM, intact & 21.29 & 4.58 & 2.0255 & 12.46 \\
Document-trained LLM, swap & 7.15 & 1.23 & 3.0156 & 11.10 \\
Document-trained LLM, mean & 10.30 & 1.97 & 2.7545 & 10.04 \\
Mean-trained LLM, intact & 16.32 & 3.73 & 2.2485 & 14.48 \\
Mean-trained LLM, swap & 12.13 & 2.50 & 2.4809 & 13.62 \\
Mean-trained LLM, mean & 12.42 & 2.78 & 2.4329 & 12.23 \\
\bottomrule
\end{tabular}
\caption{All nine absolute conditions on the same 979 questions. Adapted-LLM rows average ten fits. Intact uses the original projected document vector, swap uses a fixed different-paper vector, and mean uses the public projection of the training-only mean embedding. EM, NLL and length are descriptive. Lower swap scores for the document-trained LLM are retained.}
\label{tab:fixed-reader-absolute}
\end{table*}

\paragraph{Answer improvements, regressions and direct comparison.}
Relative to the public decoder, the document-trained decoder gains exact matches on a mean 36.2 questions per fit and loses them on 2.4, with respective ranges 25--44/1--4. It gains token overlap on 81.7 formerly zero-F1 answers, while 228.8 answers with some overlap fall to zero. Mean answer length falls from 48.90 to 12.46 tokens. Token-overlap transitions are not semantic correctness labels. With donor input, F1 falls from 10.90 for the public decoder to 7.15 after document training. Consequently, the interaction between training and document input exceeds the actual 7.08-point improvement on correct documents.

When both trained decoders receive the same correct document vector, the document-trained decoder exceeds the mean-trained decoder by $+4.975\;[3.933,6.023]$ F1 points, with crossed interval $[3.719,6.230]$. The EM difference is $+0.848\;[0.148,1.563]$ points, but its crossed interval is $[-0.035,1.731]$. Truncating predictions to the accepted answer's word length before computing F1 still favors the document-trained decoder by $5.278\;[4.105,6.471]$. This diagnostic requires gold-answer length, so it is neither deployable nor a causal correction for answer format. Neither this difference nor the mean-trained decoder's gain identifies a causal share of the overall improvement. The result shows that some answer errors can be reduced without changing compressed storage when relevant evidence is supplied. It does not establish complete information retention or a mechanism shared by all memory systems. Supplementary Sections~S9 and~S24 report complete per-seed transitions and exploratory EM/prefix families.

\section{Released-Model QA Audit}
\label{app:native-xrag}\label{app:native-diagnostics}
\paragraph{Generation with batch size one.}
We evaluate released \texttt{Hannibal046/xrag-7b} with SFR-Embedding-Mistral and repository commit \texttt{121fa4180a8c}, without training or checkpoint selection. The test contains 11,313 TriviaQA questions, each paired with its first listed ColBERTv2 passage. Metadata ranks 11,262 of these passages first and 51 second. We retain the upstream prompts, 180-token encoding, BF16 A6000 inference, greedy 100-token generation and stopping at newline, period or comma, using batch size one. Following the released pipeline does not guarantee that these records and scores match the original paper's main table.

The closed-book condition supplies the question without a vector. Full text supplies the passage as text. The document and question conditions instead encode the corresponding text into the vector slot, while zeroing sets the projected vector to zero. Three donor conditions use vectors from different passage IDs, selected by question-key hashes 6171/6172/6173 without inspecting answers or outcomes. Different passage IDs do not guarantee that the text is semantically irrelevant. Match measures whether the normalized prediction contains an accepted answer as a sequence of tokens; it differs from strict EM and token-overlap F1. Boost measures the fraction of closed-book errors corrected, and resilience measures the fraction of closed-book successes retained. Donor effects average the scores from all three assignments within each question.

\begin{table}[t]
\centering\small
\setlength{\tabcolsep}{4pt}
\begin{tabular}{@{}lrrrr@{}}
\toprule
Input & Match (\%) & EM (\%) & F1 & Tokens \\
\midrule
Closed-book & 51.59 & 30.43 & 43.54 & 10.83 \\
Full text & 61.27 & 35.80 & 50.26 & 10.75 \\
Document input & 63.03 & 48.73 & 60.15 & 6.92 \\
Question input & 59.04 & 30.36 & 46.05 & 6.35 \\
Zero vector & 52.15 & 29.22 & 42.81 & 10.99 \\
Donor 1 & 53.98 & 41.68 & 51.84 & 7.42 \\
Donor 2 & 54.04 & 41.32 & 51.72 & 7.45 \\
Donor 3 & 54.07 & 41.15 & 51.78 & 7.44 \\
\bottomrule
\end{tabular}
\caption{Released xRAG on all 11,313 repository TriviaQA questions. Match is token-substring answer inclusion; strict EM and F1 are descriptive secondary scores. No weights are fitted in this audit.}
\label{tab:native-absolute}
\end{table}

\begin{table}[t]
\centering\small
\setlength{\tabcolsep}{4pt}
\begin{tabular}{@{}lrc@{}}
\toprule
Effect & Estimate & Simultaneous 95\% CI \\
\midrule
Document $-$ closed & $+11.45$ & $[+10.30,+12.59]$ \\
Question $-$ closed & $+7.45$ & $[+6.46,+8.44]$ \\
Document $-$ question & $+4.00$ & $[+2.98,+5.01]$ \\
Document $-$ mean donor & $+9.00$ & $[+8.05,+9.96]$ \\
Mean donor $-$ closed & $+2.44$ & $[+1.59,+3.30]$ \\
Full text $-$ closed & $+9.68$ & $[+8.37,+10.99]$ \\
Zero $-$ closed & $+0.57$ & $[-0.05,+1.18]$ \\
Question boost rate & $+24.27$ & $[+22.72,+25.81]$ \\
\bottomrule
\end{tabular}
\caption{All eight released-model effects. The first seven are Match differences in percentage points; boost is a percentage conditional on closed-book errors. One maximum-standardized family uses 50,000 paired original-passage-cluster draws and a two-point benchmark.}
\label{tab:native-effects}
\end{table}

The eight conditions produce 90,504 answers. To estimate uncertainty for this fixed checkpoint and donor bank, we resample 10,984 original-passage clusters for 50,000 draws and adjust the eight-effect family using maximum-standardized deviations. The primary contrast compares question-vector input with closed-book input on Match. Its positive gain does not extend to strict EM. In a separate exploratory family, donor input exceeds closed-book EM by $10.958\;[10.250,11.666]$ points, and document input exceeds donor EM by $7.348\;[6.635,8.063]$. Encoding the question changes the computation applied to information already visible to the decoder; it is not the separately trained constant-input control used to define $C$. The original paper includes a LoRA control trained on the task data, but does not specify its exact source access. Supplementary Section~S10 retains transition rates, donor outcomes, metric comparisons and the original table's numbers. These comparisons show how the chosen comparator affects interpretation; they neither establish that the source paper omitted controls nor provide an independent test set.

\paragraph{Output length and prefix check.}
Mean output lengths in Table~\ref{tab:native-absolute} come from all saved answers. An oracle-prefix score normalizes each answer and alias, truncates the prediction to the alias's word length, computes token F1, and maximizes over aliases. All four contrasts below remain positive; this particular length adjustment therefore does not remove the pattern. Gold length is unavailable in deployment, and order/phrasing still matter. It neither estimates a causal format share nor validates every donor as semantically unrelated.

\begin{table*}[t]
\centering\small
\begin{tabular}{@{}lrrr@{}}
\toprule
Comparison & Estimate & Fixed-fit 95\% & Adjusted sensitivity 95\% \\
\midrule
Question minus closed prefix F1 & $+1.773$ & $[+0.916,+2.633]$ & $[+0.707,+2.838]$ \\
Document minus closed prefix F1 & $+15.331$ & $[+14.505,+16.161]$ & $[+14.301,+16.360]$ \\
Donor minus closed prefix F1 & $+6.575$ & $[+5.926,+7.227]$ & $[+5.766,+7.384]$ \\
Document minus donor prefix F1 & $+8.756$ & $[+8.063,+9.445]$ & $[+7.897,+9.614]$ \\
\bottomrule
\end{tabular}
\caption{Released-model oracle-prefix diagnostics. F1 uses points. All four rows form one separate follow-up family.}
\end{table*}

A fresh run of the unmodified command-line evaluator at batch size one exactly reproduces all 11,313 saved raw answers. Changing to batch size four preserves every prompt token ID but changes 9,905 answers. Match becomes 56.96\%, strict EM 1.29\%, F1 19.39 and mean answer length 27.33 tokens, compared with 51.59\%, 30.43\%, 43.54 and 10.83 at batch one. These runs use PyTorch 2.5.1/Transformers 4.48.2. Appendix~\ref{app:native-all48} separately reports batch-four evaluation with the source-specified software versions. The released stopping rule waits for every row in a batch; padding may also affect generation, so the comparison does not isolate a single cause. The source paper calls its 57.08 closed-book score EM, whereas the released evaluator selects Match. Without original per-question outputs and the exact configuration, a nearby aggregate score cannot establish identical reproduction. This subsection's contrasts use batch one; the main text uses batch four with the specified software versions.

\section{Lexical Content: Methods and Complete Primary Effects}
\label{app:xrag-content}\label{app:xrag-interface-control}
\paragraph{Inputs and fitting.}
This study asks whether a compressed earlier passage helps predict a repeated word beyond what the visible suffix or a constant vector provides. All three conditions train the public 33.56M-parameter projector, while the SFR encoder and decoder remain frozen. The projector receives the earlier document, the final 281-token suffix, or the mean of 8,192 training-prefix embeddings. In every condition, the decoder receives one projected vector and the same suffix text. Each condition independently screens the three learning rates used in the QA study for 128 updates on 512 development items, with evaluations at steps 0/32/64/128 and zero weight decay. All select $2\times10^{-4}$. Ten training-order seeds pair initialization and example permutations. Final fits use batch 64 for 512 updates and select among checkpoints 0/128/256/384/512. The PG-19 test, shared with earlier comparisons, contains 3,044 targets from 401 books. NLL is 4.8254/4.8145/4.5088 for local, constant and remote input; corresponding accuracy is 18.11/18.15/20.28\%.

\paragraph{Paired word substitutions in new articles.}
To test whether predictions respond to earlier document content while the visible suffix stays fixed, we select 217 pairs of repeated capitalized words from 67 NASA articles published in 2025, with at most four pairs per article. Selection does not inspect model predictions. Earlier mentions lie at least 512 tokens away, and inputs contain at most 2,048 tokens. In a changed version of each example, we replace selected earlier mentions and the target with a word of the same token length. The replacement's first token differs from the original and does not occur in the visible input. Hash-based selection does not match word frequency, part of speech or semantic plausibility. The final 281-token suffix remains byte-identical. These are automatically constructed word-prediction probes, rather than human-annotated factual QA. Recent publication alone does not establish absence from model pretraining.

Let $y_o,y_c$ be the original and replacement target tokens, and let $p_o,p_c$ denote model probabilities under the corresponding document versions. We measure whether changing the remote text shifts preference toward the matching target using $(m_o+m_c)/2$, where $m_o=\log p_o(y_o)-\log p_o(y_c)$ and $m_c=\log p_c(y_c)-\log p_c(y_o)$. Preference is correct in both versions when $m_o>0,m_c>0$. This pairwise criterion differs from selecting the correct token from the full vocabulary. Local and constant inputs remain identical across versions. The following tables report original and substituted scores, all four PG-19 effects and all eighteen article effects. For original articles, crossed intervals for $C/U$ include zero. After substitution, the mean-replacement contrast and several accuracy contrasts also include zero. Full-context accuracy falls from 66.36\% to 17.05\%, showing that substitution makes the prediction task harder even with the full prefix.

\begin{table*}[t]
\centering
\small
\begin{tabular}{llrr}
\toprule
Panel & Input & NLL & Accuracy (\%) \\
\midrule
Original & Local input & 2.7986 & 49.26 \\
Original & Constant input & 2.8214 & 50.41 \\
Original & Remote input & 2.6116 & 47.56 \\
Original & Full-context reference & 1.5019 & 66.36 \\
Substituted & Local input & 13.3199 & 0.00 \\
Substituted & Constant input & 13.0405 & 0.00 \\
Substituted & Remote input & 12.2926 & 0.23 \\
Substituted & Full-context reference & 8.2073 & 17.05 \\
\bottomrule
\end{tabular}
\caption{New-article first-token prediction. The three trained arms average ten paired training-order seeds. The full-context row is one frozen decoder receiving the whole prefix, not a local reference or a ten-seed fit.}
\label{tab:xrag-content-absolute}
\end{table*}

\begin{table*}[t]\centering\small
\setlength{\tabcolsep}{3pt}
\begin{tabular}{@{}lrr@{}}\toprule
Study / effect (estimate) & Fixed-fit percentile 95\% & Crossed simultaneous 95\% \\\midrule
\multicolumn{3}{l}{{Ten-fit PG-19 lexical study}} \\
$C$ (NLL) (+0.3057) & $[+0.2530,+0.3585]$ & $[+0.2369,+0.3746]$ \\
$C$ (accuracy pp) (+2.1255) & $[+1.1501,+3.0933]$ & $[+0.8538,+3.3959]$ \\
Constant-reference utility (NLL) (+0.0109) & $[-0.0042,+0.0259]$ & $[-0.0095,+0.0318]$ \\
Constant zeroing (NLL) (+1.2360) & $[+1.1719,+1.3010]$ & $[+1.1544,+1.3180]$ \\
\multicolumn{3}{l}{{NASA article lexical study}} \\
Original: $U$ (NLL) (+0.1871) & $[+0.0622,+0.3164]$ & $[-0.0038,+0.3896]$ \\
Original: $C$ (NLL) (+0.2098) & $[+0.0690,+0.3567]$ & $[-0.0058,+0.4424]$ \\
Original: $U$ (accuracy pp) (-1.7051) & $[-4.8837,+1.1345]$ & $[-7.0616,+2.9057]$ \\
Original: $C$ (accuracy pp) (-2.8571) & $[-6.5072,+0.6818]$ & $[-8.8456,+2.6953]$ \\
Original: $D_{\rm zero}$ (NLL) (+0.5873) & $[+0.3156,+0.8673]$ & $[+0.1809,+1.0253]$ \\
Original: $D_{\rm swap}$ (NLL) (+0.6074) & $[+0.4395,+0.7839]$ & $[+0.3543,+0.8844]$ \\
Original: Mean-replacement damage (NLL) (+0.7950) & $[+0.5511,+1.0591]$ & $[+0.4220,+1.2229]$ \\
Original: Constant zeroing (NLL) (+0.3775) & $[+0.1557,+0.6031]$ & $[+0.0434,+0.7260]$ \\
Substituted: $U$ (NLL) (+1.0273) & $[+0.8252,+1.2439]$ & $[+0.7097,+1.3728]$ \\
Substituted: $C$ (NLL) (+0.7479) & $[+0.5453,+0.9628]$ & $[+0.3966,+1.1293]$ \\
Substituted: $U$ (accuracy pp) (+0.2304) & $[+0.0000,+0.7317]$ & $[+0.0000,+1.4414]$ \\
Substituted: $C$ (accuracy pp) (+0.2304) & $[+0.0000,+0.7317]$ & $[+0.0000,+1.4414]$ \\
Substituted: $D_{\rm zero}$ (NLL) (+3.4998) & $[+3.1758,+3.8268]$ & $[+2.9912,+4.0108]$ \\
Substituted: $D_{\rm swap}$ (NLL) (+0.7817) & $[+0.5726,+1.0072]$ & $[+0.4615,+1.1385]$ \\
Substituted: Mean-replacement damage (NLL) (+0.1919) & $[-0.0486,+0.4355]$ & $[-0.1814,+0.5908]$ \\
Substituted: Constant zeroing (NLL) (+2.7519) & $[+2.4653,+3.0480]$ & $[+2.3136,+3.2180]$ \\
Paired: Remote log-odds response (+0.6655) & $[+0.5326,+0.8133]$ & $[+0.4686,+0.9118]$ \\
Paired: Both versions correct (pp) (+4.2396) & $[+2.1267,+6.5877]$ & $[+1.1538,+8.0375]$ \\
\bottomrule\end{tabular}
\caption{All four PG-19 and eighteen NASA effects, in separate families. Fixed-fit intervals resample documents; crossed intervals also resample ten paired order seeds with family adjustment. Original NASA $U/C$ have positive fixed-fit lower bounds and negative crossed lower bounds. NLL uses nats and accuracy uses percentage points.}
\label{tab:xrag-content-historical-pg19}\label{tab:xrag-content-fresh-nasa}
\end{table*}

\begin{table}[t]
\centering\footnotesize
\setlength{\tabcolsep}{3pt}
\begin{tabular}{@{}lrrrr@{}}\toprule
Task / score & $D_{\rm doc}$ & $C$ & $D_{\rm const}$ & Ratio \\\midrule
QASPER (F1) & 13.16 & 7.43 & 5.73 & 43.5\% \\
PG-19 (NLL) & 1.54 & 0.31 & 1.24 & 80.2\% \\
NASA original (NLL) & 0.59 & 0.21 & 0.38 & 64.3\% \\
\bottomrule\end{tabular}
\caption{Shared-zero arithmetic: $D_{\rm doc}=C+D_{\rm const}$. Ratio is $D_{\rm const}/D_{\rm doc}$, a descriptive score ratio, not a causal fraction of stored information. Units differ across rows; complete intervals are in Appendices~\ref{app:gold-qa} and~\ref{app:xrag-content}.}
\label{tab:shared-zero-summary}
\end{table}

\section{Other Comparisons and Boundaries}
\label{app:training}\label{app:joint-attention}\label{app:memory-pairs}
\label{app:external-du}\label{sec:external-du}\label{app:oa-reference}
\label{app:qa-routing}\label{supp:retrieval}\label{app:reference-audit}
\paragraph{Equal development searches and longer training.}
A separate pooled/query-memory comparison starts from the same public checkpoint and gives each condition the same development-search opportunity. All intervention intervals lie within $\pm0.01$ nat. No evaluated cohort establishes both positive zeroing damage and negative utility. In another experiment, a query reader trained on frozen representations has $U=-0.0855,D=1.343$ nats. Its PG-19 local comparator worsens on development data, so a positive gain over that comparator's final fit does not establish superiority to a well-selected no-memory model. A further study trains memory/comparator pairs for 2,048 updates, selects checkpoints on development data, and evaluates new synthetic stories and MuSiQue. All four NLL utility intervals include zero. Synthetic pooled-memory zeroing damage is $0.0278\;[0.0111,0.0446]$. On MuSiQue, zeroing query memory instead improves NLL, with $D=-0.0230\;[-0.0388,-0.0079]$. Supplementary Sections~S4--S5 provide the separate recipes and complete settings.

\begin{table*}[!t]
\centering\small
\setlength{\tabcolsep}{4pt}
\begin{tabular}{@{}lr rr cc@{}}
\toprule
System & Targets & Local NLL & Intact NLL & Gain $U$ [simultaneous CI] & Zeroing $D$ [simultaneous CI] \\
\midrule
\multicolumn{6}{@{}l}{\textit{Synthetic: state-disambiguating token}} \\
StreamingLLM & 1,652 & 1.849 & 2.459 & $-0.610\;[-0.667,-0.552]$ & $+6.484\;[+6.279,+6.694]$ \\
H$_2$O & 1,652 & 1.849 & 2.068 & $-0.219\;[-0.292,-0.147]$ & $+7.690\;[+7.467,+7.916]$ \\
xRAG & 2,000 & 2.279 & 3.211 & $-0.932\;[-1.010,-0.855]$ & $+6.790\;[+6.582,+6.995]$ \\
LoCoCo & 2,000 & 2.232 & 3.646 & $-1.414\;[-1.598,-1.234]$ & $+6.367\;[+6.150,+6.591]$ \\
CCM & 2,000 & 2.231 & 0.078 & $+2.153\;[+2.104,+2.197]$ & $+9.910\;[+9.728,+10.089]$ \\
ICAE & 1,976 & 2.244 & 0.058 & $+2.185\;[+2.139,+2.227]$ & $+7.767\;[+7.633,+7.907]$ \\
Beacon & 1,201 & 1.921 & 0.056 & $+1.865\;[+1.801,+1.926]$ & $+7.955\;[+7.756,+8.152]$ \\
MemoryLLM & 1,990 & 1.876 & 0.029 & $+1.847\;[+1.790,+1.900]$ & $+9.822\;[+9.696,+9.945]$ \\
M+ & 1,990 & 1.872 & 0.068 & $+1.804\;[+1.751,+1.858]$ & $+8.834\;[+8.707,+8.965]$ \\
\midrule
\multicolumn{6}{@{}l}{\textit{PG-19: repeated-word first token}} \\
StreamingLLM & 3,044 & 6.919 & 6.586 & $+0.333\;[+0.141,+0.530]$ & $+4.895\;[+4.576,+5.210]$ \\
H$_2$O & 3,044 & 6.919 & 2.359 & $+4.560\;[+4.353,+4.766]$ & $+8.726\;[+8.503,+8.948]$ \\
xRAG & 3,044 & 4.712 & 4.813 & $-0.100\;[-0.192,-0.006]$ & $+1.237\;[+1.103,+1.372]$ \\
LoCoCo & 3,044 & 4.488 & 3.812 & $+0.676\;[+0.516,+0.842]$ & $+5.653\;[+5.408,+5.896]$ \\
CCM & 3,044 & 4.708 & 4.246 & $+0.462\;[+0.342,+0.590]$ & $+2.200\;[+2.023,+2.382]$ \\
ICAE & 3,044 & 8.665 & 11.597 & $-2.932\;[-3.327,-2.556]$ & $-2.326\;[-2.708,-1.957]$ \\
Beacon & 3,044 & 5.416 & 2.555 & $+2.861\;[+2.631,+3.090]$ & $+4.650\;[+4.414,+4.883]$ \\
MemoryLLM & 3,044 & 4.714 & 2.202 & $+2.513\;[+2.275,+2.762]$ & $+8.106\;[+7.882,+8.324]$ \\
M+ & 3,044 & 4.754 & 8.336 & $-3.583\;[-3.807,-3.366]$ & $+1.782\;[+1.695,+1.872]$ \\

\bottomrule
\end{tabular}
\caption{Existing systems, three fitted seeds per row, in nats. Positive $U$ favors memory; positive $D$ means zeroing harms prediction. Synthetic rows score state-disambiguating tokens; PG-19 rows score repeated-word first tokens. Intervals are study-specific simultaneous 95\% intervals. H$_2$O uses full-prefix prefill followed by pruning.}
\label{tab:core-external-bounds}
\end{table*}

\begin{table*}[!t]
\centering\small
\setlength{\tabcolsep}{5pt}
\begin{tabular}{@{}lrrcc@{}}
\toprule
System & Absent NLL & Zeroed NLL & $D_{\rm abs}$ [simultaneous CI] & $E$ [simultaneous CI] \\
\midrule
\multicolumn{5}{@{}l}{\textit{Synthetic: state-disambiguating token}} \\
StreamingLLM & 10.906 & 8.943 & $+8.447\;[+8.307,+8.588]$ & $+0.002\;[-0.004,+0.005]$ \\
H$_2$O & 10.906 & 9.757 & $+8.838\;[+8.684,+8.991]$ & $+0.405\;[+0.353,+0.460]$ \\
xRAG & 9.971 & 10.001 & $+6.760\;[+6.556,+6.962]$ & $+1.268\;[+1.216,+1.321]$ \\
\midrule
\multicolumn{5}{@{}l}{\textit{PG-19: repeated-word first token}} \\
StreamingLLM & 11.582 & 11.481 & $+4.996\;[+4.687,+5.312]$ & $+0.002\;[-0.003,+0.010]$ \\
H$_2$O & 11.582 & 11.086 & $+9.223\;[+8.974,+9.476]$ & $+3.711\;[+3.463,+3.959]$ \\
xRAG & 6.012 & 6.050 & $+1.199\;[+1.067,+1.333]$ & $+0.569\;[+0.506,+0.635]$ \\

\bottomrule
\end{tabular}
\caption{Additional interventions for StreamingLLM, H$_2$O, and xRAG in Table~\ref{tab:core-external-bounds} (nats). Positive $D_{\rm abs}$ means absence harms prediction; positive $E$ means evidence masking harms more than matched sham masking. The respective simultaneous families contain 8/8/32 effects.}
\label{tab:core-evidence-sham}
\end{table*}

\paragraph{Nine systems, task rankings and average prediction.}
Supplementary Tables~S8, S9 and~S13 report all nine systems on both data sources, including every effect direction, interval and selection record. Because systems differ in which components are adapted and which inputs they can access, their results do not isolate architectural causes. CCM, Beacon and MemoryLLM outperform their trained local comparators on both sources. Four synthetic configurations worsen when memory is zeroed but still underperform their local comparators. Sensitivity to relevant evidence can therefore coexist with negative utility. In PG-19, every direct comparison between gains on all tokens and gains on repeated-word targets has an interval containing zero. Although model rankings change with the evaluated task, cross-fitted comparisons do not establish an advantage from selecting models by memory-target scores. These results bound the interpretations of the observed counterexamples.

\paragraph{Trained constant projector versus directly optimized vector.}
With a fixed decoder, a biased projector receiving one constant input and a directly optimized 4,096-dimensional vector can represent the same set of constant outputs. Finite training need not find the same output. We optimize the free vector using gold-answer NLL, rather than a KL objective that preserves a model's outputs. One comparison keeps the decoder frozen; another jointly adapts rank-16/alpha-32 decoder LoRA. Each uses ten paired fits, 512 updates, batch 16 and QA development selection. Both joint conditions select learning rate $5\times10^{-5}$; both free-vector conditions select coefficient $0.1$, the largest tested multiplier of the root mean square of the public projected mean. Warm starts use the trained constant projector's output and therefore include prior training. Let $G=S_{\rm trained\ constant}-S_{\rm vector}$ be the score difference between the two forms of constant input. The change in that gap under joint decoder adaptation is F1 $H=G_{\rm joint}-G_{\rm frozen}=0.449\;[-0.106,1.010]$, with crossed interval $[-0.650,1.548]$. For NLL reduction, $H=0.0283\;[0.0207,0.0358]$, with crossed interval $[0.0111,0.0454]$. Thus joint adaptation does not establish a larger F1 difference between these controls. The likelihood result also does not demonstrate an error in the optimal-ablation method, whose objective differs from ours. Supplementary Section~S11 retains all fourteen effects and selections.

\paragraph{Choosing when to use a document and evaluating retrieved text.}
We train question-only rules to decide which questions receive document input under a fixed usage budget. Four rules predict, respectively, the gain over trained constant input, swap damage, zeroing damage, or the gain over trained question input. All use the same fixed features, centered linear-kernel ridge model and five-fold cross-validation with whole papers kept together. At 489/979 document uses, the rule trained on $C$ exceeds the swap-trained rule by $0.375\;[-0.133,0.921]$ F1 points. Its direct differences from the zeroing-trained and question-comparison rules also lie within $\pm2$. Its gain over uniform allocation is $1.632\;[1.101,2.187]$, which does not establish an advantage larger than two points. A separate input-shift diagnostic replaces annotated evidence with the top-1 or top-3 paragraphs retrieved within the question's paper by TF--IDF or SFR. It does not retrain the model on retrieved text. No retrieval policy meets the specified criterion of negative gains over both trained controls together with positive intervention damage beyond the stated magnitude benchmarks. Supplementary Sections~S21--S23 give that criterion, complete rules, budget curves, all retrieval policies and both interval constructions.

\begin{table*}[t]\centering\small
\begin{tabular}{@{}lrrrr@{}}\toprule
Policy & Intact F1 & $U_{\rm question}$ & $C$ [crossed 95\%] & $D_{\rm swap}-C$ [crossed 95\%] \\\midrule
TF--IDF top-1 & 13.213 & $+0.678$ & $+0.668\;[-1.369,2.706]$ & $+5.818\;[4.015,7.621]$ \\
TF--IDF top-3 & 14.050 & $+1.515$ & $+1.505\;[-0.145,3.154]$ & $+5.217\;[3.519,6.915]$ \\
SFR top-1 & 15.597 & $+3.062$ & $+3.052\;[0.948,5.155]$ & $+5.962\;[4.222,7.702]$ \\
SFR top-3 & 15.880 & $+3.345$ & $+3.335\;[1.375,5.295]$ & $+5.272\;[3.545,6.999]$ \\
\bottomrule\end{tabular}
\caption{Retrieved-input deployment shift on the same 979 questions. Shared question/constant F1 is 12.535/12.545. Bounds belong to the full 28-effect seed--paper simultaneous family. All four policies have damage exceeding $C$, without establishing negative utility or $U/C$ divergence. All 28 effects and absolute EM/NLL/length remain in the companion.}
\label{tab:core-retrieval}\end{table*}

\paragraph{Scores normalized by closed-book and full-text models.}
The ratio $(F_a-F_{\rm closed})/(F_{\rm full}-F_{\rm closed})$ expresses a condition's F1 gain over closed-book input as a fraction of the gap between full-text and closed-book F1. Here $F_a$ is the condition being evaluated. On our QA outputs, this ratio is 25.87\% for the trained constant and 58.24\% for document input. We recompute the denominator in every bootstrap draw. A positive normalized score can therefore occur even without document access. These ratios neither measure recovered information nor replicate results with task-finetuned teacher models. Across 722/9,790 seed--question pairs (7.37\%), document input beats the frozen closed-book model but loses to the trained constant. Ties remain in the denominator. Supplementary Section~S8.1 reports all eight effects and the reverse transitions.

\section{Inference, Scope and Reproduction}
\label{app:statistics}\label{supp:interval-sensitivity}\label{app:analysis-scope}
\paragraph{Paired uncertainty.}
Fixed-fit document intervals treat the trained models as fixed and capture variation across evaluation sources. We first average paired seeds for each item, then resample complete stories, books or papers with replacement. Each draw recomputes the specified item-weighted mean; its 2.5th/97.5th percentiles give an unadjusted 95\% interval. Related targets and original/substituted versions stay together. QA, decoder and lexical analyses use 50,000 draws. The released-model audit at batch one clusters questions by retrieved passage. Batch-four evaluations instead join questions sharing either a batch or a passage and resample those connected components. Multiple-comparison adjustments remain specific to each stated study family, even when results share a table.

Crossed intervals additionally resample paired training-order seeds, capturing this source of training variation as well as evaluation-source variation. QA, reference-score, allocation, optimal-ablation and decoder families use maximum-standardized deviations for simultaneous coverage. Lexical analyses and the specified fixed-fit families use Bonferroni percentiles $0.025/m$ and $1-0.025/m$, where $m$ is the number of effects in the family. Tables~\ref{tab:qa-core-sensitivity}, \ref{tab:fixed-reader-effects}, \ref{tab:xrag-content-fresh-nasa} and \ref{tab:core-retrieval} provide both constructions or the corresponding crossed bounds for the core claims; Supplementary Section~S26 gives additional diagnostics. The two constructions cover different sources of uncertainty. A lower bound above zero supports the direction of a gain; a lower bound above a numerical benchmark supports a gain exceeding that magnitude. An interval within the positive and negative benchmarks describes its range but does not establish practical equivalence. The stated benchmarks are 0.01 nat, two F1 points or one accuracy point. They are not validated deployment thresholds. For the decoder study's joint claim, every primary lower bound must exceed zero.

\begin{table*}[t]\centering\small
\begin{tabular}{@{}p{.29\textwidth}p{.27\textwidth}p{.37\textwidth}@{}}\toprule
Comparison set & Adjusted family & Inference scope \\\midrule
QA projectors & 12 F1/NLL effects & Ten paired order seeds and papers \\
Fixed-storage LLMs & 7 original F1; 2 direct F1/EM & Each family paired over seeds and papers \\
Released QA model & 8 Match/boost effects & One checkpoint; passage clusters \\
Lexical projectors & 4 PG-19; 18 NASA & Separate tasks; paired seeds/documents \\
Optimized/trained constants & 14 F1/NLL effects & Sample and interval choices use available results \\
Allocation / retrieval & 6 / 28 F1 effects & Rule quality versus deployed inputs \\
Secondary EM / prefix & EM 4/3/2; prefix 2/4 & Saved-output follow-ups \\
Planned comparisons & 6 contrasts & Paired seed/source or passage clusters \\
Pinned generation & 15 score contrasts & Batch/passage connected components \\
Additional pinned datasets & 12 score contrasts & Two datasets; batch/passage components \\
CoQA transfer & 3 F1 contrasts & Paired order seeds and stories \\
\bottomrule\end{tabular}
\caption{Historical study families, supplemented by the six planned contrasts, fifteen pinned-generation effects and three CoQA effects in Appendix~\ref{app:followup-families}. Secondary analyses are exploratory, not independent confirmations. No bound accounts for selection among analyses. Other studies retain their original fixed-fit adjustments; the numerical artifact lists all family members.}
\label{tab:family-map}\end{table*}

\paragraph{Selection and scope.}
The intervals do not cover arbitrary initialization choices, searches over configurations or selection among analyses. Analyses of normalized scores, document allocation, slot interventions, local candidates, task rankings, cross-fitted model selection and joint effects use available test summaries and are exploratory. Each training study retains its stated development-selection or fixed-final-checkpoint rule. In the ten-fit constant-input study, available results informed both sample size and the displayed interval specification. Repeated analysis of the same public test is not independent confirmation. Crossed intervals capture variation in training order at the shared initialization and recipe. The main Limitations and the complete analysis-scope table apply to every result.

\paragraph{Complete evidence and files.}
This appendix contains core implementation details and all principal objective/content, QA, LLM and lexical contrast families. It also includes the nine-system bounds, retrieval summary, native lengths and prefix contrasts. The separate \texttt{supporting\_evidence.pdf} preserves the full settings, additional-system studies, all per-seed transitions, development curves, rankings, allocation/retrieval diagnostics, exact-match/prefix follow-ups and complete interval tables. Sections/tables prefixed S refer to that document. No outcome is removed by sign.

The source/data snapshot retains saved answers and scores, fixed input identities, fitting/selection records, analysis sources and both PDF sources. Its standalone CoQA reconstruction officially rescores 20,250 saved answers and reconstructs the three paired effect intervals; it does not rerun neural inference. Earlier portable checks covering 196 historical effects are not a coverage claim for the expanded studies. Model weights are excluded, and full model reproduction requires the recorded original inputs and runtime dependencies. The numerical archive preserves each analysis family and every retained unfavorable result.

\section{Review of Controls in Selected Memory Papers}
\label{app:literature-audit}
We examined the full texts and appendices of 40 selected papers: 19 on learned compression and 21 on attention key/value (KV) retention. We fixed the paper list and seven categories before classification, recorded source URLs and PDF hashes, and kept page-level evidence for each judgment. This is a purposive review of named papers, not a systematic or random census.

\paragraph{What we looked for.}
A trained independent input supplies a learned constant or learned tokens through the tested interface without using the current document; we identify controls that replace only part of a memory. We distinguish this from training a model without access to the source, and from removing memory after training. Other categories record full-input comparisons, frozen source-restricted models, broader task-trained comparisons, changes to source content or order, and tasks that score source recall, reconstruction or answers. ``Not located'' means we found no qualifying comparison in the reviewed version; ``unclear'' means the description did not establish the required input access or interface.

\begin{table*}[t]\centering\small
\begin{tabular}{@{}lrrrrrrrr@{}}\toprule
 & \multicolumn{4}{c}{Learned compression ($n=19$)} & \multicolumn{4}{c}{KV policy ($n=21$)} \\
Category & Present & Not located & Unclear & n/a & Present & Not located & Unclear & n/a \\\midrule
Full input/cache & 16 & 3 & 0 & 0 & 21 & 0 & 0 & 0 \\
Frozen source restriction & 4 & 12 & 1 & 2 & 3 & 18 & 0 & 0 \\
Trained reference (broad) & 16 & 3 & 0 & 0 & 3 & 0 & 0 & 18 \\
Trained independent input & 0 & 17 & 2 & 0 & 1 & 20 & 0 & 0 \\
Source/order intervention & 2 & 17 & 0 & 0 & 3 & 18 & 0 & 0 \\
Same-fit zero/removal & 4 & 12 & 3 & 0 & 7 & 14 & 0 & 0 \\
Direct content task & 13 & 6 & 0 & 0 & 15 & 6 & 0 & 0 \\
\bottomrule\end{tabular}
\caption{Sample descriptions under the stated broad definitions. Unclear and not-applicable cases are shown, not folded into absence. ``Present'' includes the qualified component/order controls described below. A missing control does not invalidate a paper's performance or efficiency claim.}
\label{tab:literature-counts}
\end{table*}

\paragraph{Findings and their scope.}
Among the 19 learned-compression papers, we found no clear instance of a trained independent input under this definition: 17 were not located and two were unclear. Gist's hidden-instruction control and RMT's nonrecurrent learned tokens are related precedents, but their equivalence to the focal compressed interface remains unclear. Among the 21 KV papers, StreamingLLM supplies a component-specific precedent by learning initial sink slots while retaining other source KV. These distinctions do not support a claim that comparable controls are absent throughout the field, or that any paper's performance claims are invalid.

We retained separate classification records for 24 papers, covering a deterministic 12-paper sample and additional ambiguous or contribution-relevant cases. The records disagreed in 18 of 168 cells; we preserved the disagreements and their resolutions. Agreement was 89.29\% with pooled nominal Cohen $\kappa=0.820$, reported descriptively rather than as validated reliability. The AI Assistance statement discloses the assistance used in the review. Supplementary Section~S34 gives all paper-level categories, source links and qualifications.

\section{Within-Paper Evidence and Matched Decoder Controls}
\label{app:samepaper-followup}
This comparison asks whether the benefit of annotated evidence remains when the replacement text comes from the same paper, rather than a different topic. We specify the contrast before generating its outputs, but reuse the previously evaluated public QASPER test; this is not independent confirmation. All 979 questions from 388 papers remain eligible. We reuse the same ten projectors trained on document input, with seeds 501--510, and keep questions, decoder prompts and generation settings fixed.

For each question, the donor text comes from that question's paper. We exclude paragraphs that match annotated evidence after whitespace normalization, including substring overlap in either direction. A question-specific hash orders the remaining paragraphs. We take a prefix whose SFR input length, including special tokens, exactly matches the original evidence input; the final paragraph may be partial. Selection inspects neither predictions nor answer matches. Text without an evidence annotation may still contain the answer. Thus matching the paper and encoder length controls topic and length differences, but does not isolate answer-bearing semantics from every formatting or content difference.

Mean intact/same-paper F1 is 19.9785/4.7420, giving an intact-minus-swap gap of 15.2365 points. With 50,000 paired whole-paper draws, the fixed-fit 95\% interval is $[13.8176,16.6779]$; crossed paired order-seed/paper resampling gives $[13.7586,16.7300]$. The crossed Bonferroni interval is $[13.2351,17.2696]$: individual nominal coverage 99.1667\%, retaining the original six-primary-contrast family for nominal 95\% simultaneous coverage. The six-comparison family is defined in Appendix~\ref{app:followup-families}. These intervals do not cover analysis-selection or configuration-search uncertainty. Saved answer text was independently rescored, and a separately indexed bootstrap agreed with the reported reconstruction.

\paragraph{Matched decoder training.}
A second comparison tests whether correct document vectors help when all decoder conditions receive the same training budget and model-selection opportunities. Ten paired training-order fits train with correct document vectors, vectors from other papers, or questions alone (closed-book). They share learning-rate and checkpoint grids, 512 updates, batch size 16 and paired training orders. Wrong-document vectors come from another paper and change across uses during training. At evaluation, both decoder conditions trained with vectors receive the same correct document vectors; the closed-book decoder still receives only the question. Development data select checkpoints, while the reused public test provides no independent confirmation. These fits are separate from the earlier document/mean decoder comparison.

Mean F1 is 21.2108 for correct-document training, 13.7265 for wrong-document training, and 13.2666 for closed-book training. Correct minus wrong-document training is +7.4844 F1; correct-document minus closed-book training is +7.9442. Both remain positive under the prespecified six-primary Bonferroni intervals. This supports document-input benefit under matched training, without establishing complete recovery of the source text.

\begin{table*}[t]
\centering\small
\begin{tabular}{lrrr}
\toprule
Contrast (F1 points) & Estimate & Crossed 95\% CI & Six-primary adjusted CI \\
\midrule
Intact minus same-paper swap & 15.2365 & $[13.7586,16.7300]$ & $[13.2351,17.2696]$ \\
Correct minus wrong document & 7.4844 & $[6.2284,8.7535]$ & $[5.7842,9.1670]$ \\
Correct document minus closed book & 7.9442 & $[6.6133,9.2603]$ & $[6.1394,9.7217]$ \\
\bottomrule
\end{tabular}
\caption{Planned QASPER contrasts. Crossed resampling pairs order seeds and whole papers; the adjustment retains the original six-contrast family.}
\end{table*}

\paragraph{Decoder and full-text training settings.}
All four arms, including the exploratory full-text reference, adapt 41,943,040 decoder parameters with rank-16, scale-32, zero-dropout LoRA; the base weights, encoder and public projector remain frozen. The 1,458 training questions and paired seeds 501--510 are shared. Each arm independently screens learning rates $10^{-5},5\times10^{-5},2\times10^{-4}$ for 128 updates at seed 491 on the first 128 development questions; all select $5\times10^{-5}$. Final fits use batch 16, 512 updates, weight decay zero, FP32 trainables with BF16 base weights, AdamW $\beta=(0.9,0.95)$, clipping one, eight warmup updates, and cosine decay to one tenth of peak. Gradient checkpointing is enabled in all four arms. Minimum development canonical-answer NLL selects among steps 0,16,32,64,128,256,384,512; ties favor the earlier step. Training scores answer tokens including EOS, averaged per question.

The full-text condition supplies the same annotated evidence as ordinary text instead of a compressed vector, retaining the question and answer prefix. Closed-book prompts omit the background; the two compressed conditions retain one vector slot. All use greedy generation, EOS stopping and a 96-token cap. Test answers never select checkpoints. The full-text decoder therefore has the same training budget and trainable parameter count, but receives a longer input. Its comparison is exploratory because it lies outside the planned six-comparison family.

\begin{table}[t]\centering\small
\begin{tabular}{ll}\toprule
Decoder & Selected update: number of seeds\\\midrule
Correct document & 64:10 \\
Wrong document & 64:10 \\
Closed book & 64:8, 128:2 \\
Full text & 64:7, 128:3 \\
\bottomrule\end{tabular}
\caption{Development-selected checkpoints for the four decoder arms; each arm still completes all 512 updates.}
\end{table}

\paragraph{Exploratory secondary results.}
The decoder trained with evidence text reaches F1 62.5427. The compressed-document minus full-text F1 difference is -41.3318 points (crossed 95\% interval $[-43.6812,-38.9475]$), so the compressed model's positive gains over other controls do not establish full recovery of the available text's predictive benefit. In the earlier decoder study, replacing evidence with text from the same paper lowers F1 by 13.2977 for the document-trained decoder and 3.4352 for the mean-trained decoder. The difference between these drops is +9.8625, with crossed 95\% interval $[8.5727,11.1766]$. These are exploratory secondary comparisons. Their nominal intervals do not adjust for selection or multiplicity across exploratory analyses.

For the two matched decoder contrasts, fixed-fit whole-paper 95\% intervals are $[6.3188,8.6545]$ and $[6.7291,9.1653]$, respectively. Raw saved answers were independently rescored and an alternative indexed bootstrap reconstructed the results. Integrity checks read the actual trainable tensors, verified changing other-paper donors and paired training order, and checked storage byte and row hashes. These checks validate recorded computations, not the broader interpretation of the contrasts.

\section{Evaluation with Fixed Software Versions and Batch Size Four}
\label{app:native-all48}
We repeat all eight input conditions on the same 11,313 TriviaQA questions and released checkpoint, using the repository's default batch size of four and its specified software versions. We choose these generation settings from the source documentation before inspecting their outputs. This defines a reproducible evaluation setting; the original per-question outputs are unavailable, so it does not establish identical predictions. Two additional diagnostics each examine 256 questions per condition to test sensitivity to ordering and batch membership. Evaluations with batch size one and with a different software runtime remain separate sensitivity analyses.

\begin{table}[t]\centering\small
\begin{tabular}{lrrrr}\toprule
Input & Match & EM & F1 & Tokens\\\midrule
closed & 57.01 & 1.31 & 19.38 & 27.35 \\
full & 65.87 & 1.64 & 21.86 & 27.82 \\
document & 65.73 & 6.33 & 29.88 & 21.17 \\
question & 60.53 & 1.66 & 22.43 & 15.53 \\
zero & 57.61 & 1.28 & 19.49 & 27.45 \\
swap1 & 56.44 & 5.19 & 24.80 & 21.94 \\
swap2 & 56.37 & 5.14 & 24.75 & 22.37 \\
swap3 & 56.54 & 5.00 & 24.78 & 21.77 \\
\bottomrule\end{tabular}
\caption{All input conditions with fixed software versions and batch size four. Closed supplies only the question; full supplies passage text; document/question supply the corresponding compressed vector; zero supplies a zero vector. Swap1--3 use vectors from different passages. Match measures answer inclusion, EM strict normalized exact match, and F1 token overlap. Donor comparisons average the three swaps.}
\end{table}

For uncertainty estimates, we join questions that share an original batch or a retrieved passage and resample the resulting 2,501 connected components. This preserves dependencies created by shared generation and source text. Five contrasts, each measured with Match, EM and F1, form a separate 15-effect family. Its adjustment does not replace any earlier primary or secondary family.

\begin{table*}[t]\centering\small
\begin{tabular}{llrrr}\toprule
Contrast & Metric & Estimate & Ordinary 95\% & 15-effect adjusted\\\midrule
donor minus closed & MATCH & -0.5657 & $[-1.1804,0.0505]$ & $[-1.5033,0.3651]$ \\
donor minus closed & EM & +3.8039 & $[3.4083,4.2064]$ & $[3.2191,4.4162]$ \\
donor minus closed & F1 & +5.4011 & $[4.8976,5.9070]$ & $[4.6438,6.1705]$ \\
document minus donor & MATCH & +9.2814 & $[8.6003,9.9693]$ & $[8.2599,10.3119]$ \\
document minus donor & EM & +1.2169 & $[0.7362,1.7047]$ & $[0.5139,1.9439]$ \\
document minus donor & F1 & +5.0961 & $[4.5498,5.6441]$ & $[4.2815,5.9208]$ \\
document minus question & MATCH & +5.1976 & $[4.4399,5.9671]$ & $[4.0698,6.3312]$ \\
document minus question & EM & +4.6672 & $[4.1047,5.2432]$ & $[3.8171,5.5331]$ \\
document minus question & F1 & +7.4435 & $[6.8257,8.0661]$ & $[6.5193,8.3899]$ \\
document minus zero & MATCH & +8.1234 & $[7.3193,8.9272]$ & $[6.9513,9.3175]$ \\
document minus zero & EM & +5.0473 & $[4.5278,5.5842]$ & $[4.2725,5.8477]$ \\
document minus zero & F1 & +10.3828 & $[9.7333,11.0406]$ & $[9.4141,11.3508]$ \\
document minus full & MATCH & -0.1414 & $[-0.8739,0.5794]$ & $[-1.2500,0.9484]$ \\
document minus full & EM & +4.6937 & $[4.1425,5.2534]$ & $[3.8665,5.5310]$ \\
document minus full & F1 & +8.0160 & $[7.3420,8.7030]$ & $[7.0060,9.0445]$ \\
\bottomrule\end{tabular}
\caption{Pinned batch-four paired score contrasts in percentage points. Bonferroni intervals jointly cover the fifteen specified contrasts nominally at 95\%; they are not a correction over every analysis in the paper.}\end{table*}

Providing an encoded question raises Match by +3.5181 points over closed-book input; providing an encoded document raises it by +8.7156. Their ratio, 40.4\%, is a post hoc descriptive quantity outside the fifteen-effect family. It is not the fraction of document benefit caused by question information: the decoder already receives the question text in both conditions, and the extra encoded vector changes its computation.

\paragraph{Batch composition.}
Reversing question order within each batch changes zero raw answers across all eight conditions. Changing which questions share a batch changes 216--232 of 256 answers per condition. Match shifts by at most 0.390625 points, while F1 changes range from -2.1225 to +2.9714. These observations describe the fixed diagnostic subset, rather than establishing invariance across the population. Shared stopping, padding and numerical generation may all contribute; the checks do not isolate answer length as a cause. On the full evaluation set, donor vectors improve EM and F1 relative to closed-book input, while the donor Match interval includes zero. Thus the positive secondary scores do not establish a donor advantage on the source paper's reported answer-inclusion metric.

\section{Objective Comparison with Equal Development Searches}\label{app:matched-objective-final}
This study asks whether a memory model's advantage over its comparator can increase even when both models' absolute prediction loss worsens. We compare a local adapter, pooled document memory and pooled memory trained with constant input. Each is trained either to predict all next tokens or to predict only the designated state targets. Three models, two objectives and ten paired training-order seeds give sixty final fits. Every model/objective combination receives the same development-search budget over three candidate learning rates. A new test contains 1,024 targets from separately generated stories; scores on the previously evaluated panel remain secondary evidence. Training permits up to 1,024 updates. Forty-three fits reach that cap without convergence; seventeen meet the development-plateau rule. Neither stopping on a plateau nor selecting a checkpoint establishes optimality. The earlier three-seed, 512-update study uses a different recipe and does not establish negative all-token utility in this study.

\paragraph{Fitting and selection.}
All cells use Llama-3.2-3B, window 11, block size 16, one-block delay, memory width 256, and rank-16/scale-32 LoRA on q/k/v and gate/up/down projections with zero dropout. The same 2,048 training targets and 128-target development panel are reused. Each arm/objective screens LR $\{5\times10^{-5},2\times10^{-4},8\times10^{-4}\}$ for 256 updates at seed 610, selecting minimum development target NLL at steps 128/256; ties favor lower LR then earlier step. Selected all-token rates for adapter/pooled/constant are $2/8/8\times10^{-4}$; target-only rates are $8/2/2\times10^{-4}$. The ten final seeds 611--620 use batch four, AdamW $\beta=(0.9,0.95)$, zero weight decay, gradient clipping one, and 16 warmup updates followed by constant LR. Trainables and the output head are FP32; backbone weights are BF16. Development is evaluated at step zero and every 128 updates. After at least 512 updates, three successive evaluations without improving the running meaningful best by 0.001 nat trigger a plateau stop; reaching 1,024 always receives the nonconverged-cap label. The minimum NLL among all visited checkpoints is selected, ties earlier. Test results never select rates, stopping or checkpoints.

\begin{table*}[t]\centering\small
\begin{tabular}{llrrrrr}\toprule
Objective & Arm & NLL & Full accuracy (\%) & Candidate accuracy (\%) & Cap / 10 & Plateau / 10\\\midrule
all-tokens & adapter & 2.3189 & 15.51 & 15.51 & 8 & 2 \\
all-tokens & pooled & 2.3190 & 14.24 & 14.24 & 9 & 1 \\
all-tokens & constant & 2.3106 & 14.93 & 14.93 & 7 & 3 \\
target & adapter & 3.2729 & 14.00 & 18.37 & 7 & 3 \\
target & pooled & 2.4921 & 19.66 & 19.66 & 6 & 4 \\
target & constant & 2.4566 & 19.82 & 19.82 & 6 & 4 \\
\bottomrule\end{tabular}\caption{New-holdout target-token scores and stopping by arm; all ten fits retained. The candidate-informed uniform reference has mean NLL 1.8871 and accuracy 17.59\%. It is not a uniform distribution over the full vocabulary.}\end{table*}

With the target-only objective, the adapter's NLL remains 2.5805 after renormalizing probabilities over the permitted candidates. This is worse than uniform prediction over those candidates, so probability assigned outside the candidate set does not fully explain its high loss. Its full-vocabulary accuracy is below the expected accuracy of uniform candidate selection, although its candidate-restricted accuracy is above that benchmark. The memory model's relative advantage therefore uses a comparator with poor absolute performance. Switching from all-token to target-only training increases adapter NLL by 0.9539 and pooled-memory NLL by 0.1730. Pooled memory gains a relative advantage because its comparator worsens more, not because its own NLL improves. Under all-token training, the pooled-memory versus adapter interval includes zero.

\begin{table*}[t]\centering\small
\begin{tabular}{llrrr}\toprule
Panel & Contrast (nats) & Estimate & Crossed 95\% & Adjusted interval\\\midrule
New holdout & Objective difference in U & +0.7809 & $[0.4656,1.0939]$ & $[0.3668,1.1963]$ \\
New holdout & All-token U & -0.0001 & $[-0.0596,0.0625]$ & $[-0.0797,0.0868]$ \\
New holdout & Target-only U & +0.7808 & $[0.4593,1.1087]$ & $[0.3611,1.2187]$ \\
New holdout & Target-only constant U & +0.8162 & $[0.5059,1.1311]$ & $[0.3619,1.2961]$ \\
New holdout & Target-only C & -0.0354 & $[-0.1101,0.0342]$ & $[-0.1418,0.0590]$ \\
Reused panel & Objective difference in U & +0.7469 & $[0.4302,1.0664]$ & $[0.3298,1.1722]$ \\
Reused panel & All-token U & +0.0167 & $[-0.0443,0.0816]$ & $[-0.0645,0.1076]$ \\
Reused panel & Target-only U & +0.7635 & $[0.4230,1.1164]$ & $[0.3203,1.2346]$ \\
Reused panel & Target-only constant U & +0.8113 & $[0.4715,1.1597]$ & $[0.3154,1.3475]$ \\
Reused panel & Target-only C & -0.0477 & $[-0.1135,0.0102]$ & $[-0.1436,0.0285]$ \\
\bottomrule\end{tabular}\caption{Paired seed/story intervals. Constant utility uses its original nineteen-secondary-effect adjustment; other displayed adjusted intervals use six-comparison endpoints, with only the new-holdout objective difference a member of the six primary contrasts. Reused-panel estimates are secondary.}\end{table*}

\begin{figure}[t]\centering
\includegraphics[width=0.94\columnwidth]{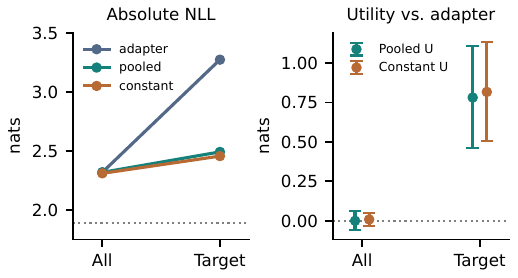}
\caption{New holdout, ten paired fits per arm/objective. Target-only training worsens target-token NLL for all arms, most for the adapter; positive utility need not imply absolute improvement. Bars: crossed seed/story 95\% intervals. Dotted NLL line: uniform-candidate reference. Appendix~\ref{app:matched-objective-final} gives adjusted intervals and accuracy.}
\label{fig:objective-control}\end{figure}

\section{CoQA Evaluation without Retraining QASPER Models}\label{app:coqa-final}
We ask whether the already trained QASPER models retain their document advantage on another QA dataset. We evaluate 405 first-turn questions from distinct CoQA development stories \citep{reddy2019coqa}, using ten paired training-order seeds and five conditions. No model is trained or selected on CoQA: all forty selected QASPER fits remain fixed, and evaluation reuses the exact memory. Official scoring averages the best token F1/EM over four human reference answers with each reference held out in turn, retaining duplicates. Inputs use annotated supporting text, called rationales. The comparison text comes from the same story outside those rationales and may still contain the answer. Encoder inputs average 24.76 tokens (median 19; range 5--197, including special tokens). This evaluates short supplied evidence, rather than full-document or multi-turn conversational QA.

This subset had not previously been evaluated in our local experiments, but CoQA training data were included in the released xRAG model's upstream instruction training. The evaluation therefore does not establish transfer to an unseen task or freedom from pretraining contamination. All CoQA conditions share one GPU and runtime configuration, while QASPER uses another. Differences in absolute scores between these datasets cannot be interpreted as hardware-controlled effects.

\begin{table}[t]\centering\small
\begin{tabular}{lr}\toprule
Fit and evaluation input & F1\\\midrule
Projector: correct rationale & 60.56\\
Projector: same-story non-rationale & 5.14\\
Correct-document decoder & 66.56\\
Wrong-document-trained decoder & 34.22\\
Closed-book decoder & 18.86\\\bottomrule\end{tabular}
\caption{Absolute CoQA F1, averaged over ten fixed QASPER fits. Decoders trained with correct and wrong document vectors both receive the same correct rationale vector at evaluation.}\end{table}

\begin{table*}[t]\centering\small
\begin{tabular}{lrrrr}\toprule
Contrast (F1) & Estimate & Fixed-fit 95\% & Crossed 95\% & Three-effect adjusted\\\midrule
Rationale minus non-rationale projector & 55.4226 & $[51.9354,58.8963]$ & $[51.8655,58.9748]$ & $[51.0388,59.7601]$\\
Correct minus wrong-document training & 32.3428 & $[28.2753,36.4576]$ & $[28.0996,36.6024]$ & $[27.2110,37.5917]$\\
Correct document minus closed book & 47.7046 & $[43.6950,51.6845]$ & $[43.5806,51.7956]$ & $[42.6608,52.6938]$\\
\bottomrule\end{tabular}\caption{Separate three-contrast family with paired story/seed resampling. Large gains are conditional on short oracle evidence and known upstream task exposure.}\end{table*}

\section{Planned Follow-up Comparisons}\label{app:followup-families}
Six contrasts were fixed after the earlier studies but before their own new outputs. The family comprises: (1) target-only minus all-token pooled utility on new synthetic holdout stories; (2) correct-evidence minus same-paper non-evidence projector F1 on QASPER; (3) correct-document minus wrong-document-trained decoder F1 on identical QASPER vectors; (4) correct-document minus equally trained closed-book decoder F1 on QASPER; (5) donor-minus-closed strict EM averaged equally across two additional repository QA datasets; and (6) document-minus-donor strict EM averaged across those same datasets. The last two use the earlier batch-one setting, not the pinned batch-four TriviaQA panel.

Each of these six two-sided intervals uses tail probability $0.025/6$ for nominal familywise 95\% coverage. Resampling pairs training-order seeds and source stories/papers for fitted models, or retrieved-passage clusters within each dataset for the frozen model. QASPER reuses a public test, and none of these bounds covers analysis selection or arbitrary initialization. The fifteen pinned-generation effects and three CoQA contrasts retain separate families; post hoc ratios and the learned full-text comparison are descriptive or exploratory, not extra primary successes.

\begin{table*}[t]\centering\footnotesize
\setlength{\tabcolsep}{3pt}
\begin{tabular}{lrrr}\toprule
Planned contrast & Estimate & Ordinary crossed/clustered 95\% & Six-contrast adjusted\\\midrule
Objective difference in utility (nats) & 0.7809 & $[0.4656,1.0939]$ & $[0.3668,1.1963]$\\
QASPER intact minus same-paper swap (F1) & 15.2365 & $[13.7586,16.7300]$ & $[13.2351,17.2696]$\\
QASPER correct minus wrong training (F1) & 7.4844 & $[6.2284,8.7535]$ & $[5.7842,9.1670]$\\
QASPER correct minus closed book (F1) & 7.9442 & $[6.6133,9.2603]$ & $[6.1394,9.7217]$\\
NQ-open/WebQuestions donor minus closed (EM) & 7.8904 & $[7.1712,8.6193]$ & $[6.9193,8.8858]$\\
NQ-open/WebQuestions document minus donor (EM) & 4.4157 & $[3.6631,5.1656]$ & $[3.4114,5.4279]$\\
\bottomrule\end{tabular}\caption{All six planned comparisons, in the order defined above. The last two average NQ-open and WebQuestions equally under batch-one generation and pair passage clusters; they are not pinned TriviaQA Match claims. Score effects are percentage points except the first row.}\end{table*}

\section{Training Configurations for the Nine-System Comparison}\label{app:nine-config}
Within each system, the memory model and its separately trained local comparator predict the same task targets; the local comparator cannot access the remote memory bank. Across systems, the adapted components, available input and training histories differ. Synthetic StreamingLLM and H$_2$O apply their cache policies to a common Llama-3.2-3B checkpoint trained with full context, retaining 384 cache positions. H$_2$O prunes after processing the full source, so this setting does not test reductions in peak memory during online input processing. xRAG trains its projector with encoder and decoder frozen; its suffix-only comparator instead trains a decoder adapter. LoCoCo trains a newly initialized convolutional compressor, while CCM adapts its released merge module. ICAE, Beacon, MemoryLLM and M+ retain their released compressors and train task decoder adapters, which are disabled while compressing the prefix. The four synthetic configurations that underperform their local comparators also have low absolute task performance. They differ in decoder adaptation from the configurations that predict well, so they do not demonstrate dependence without utility in a well-functioning memory system.

\begin{table*}[t]\centering\small
\begin{tabular}{llll}\toprule
System & Dataset & Method learning rate/decay & Local learning rate/decay\\\midrule
StreamingLLM & Syn. & 1e-3/0.1 & 1e-5/0 \\
H$_2$O & Syn. & 1e-3/0.1 & 1e-5/0 \\
xRAG & Syn. & 5e-5/0 & 5e-5/0 \\
LoCoCo & Syn. & 0.05/0 & 1e-3/0.1 \\
CCM & Syn. & 5e-05/0 & 1e-3/0 \\
ICAE & Syn. & 5e-05/0 & 5e-5/0 \\
Beacon & Syn. & 0.0002/0.1 & 2e-4/0.1 \\
MemoryLLM & Syn. & 5e-05/0 & 2e-4/0.1 \\
M+ & Syn. & 0.0002/0.1 & 2e-4/0.1 \\
\midrule
StreamingLLM & PG-19 & 5e-5/0 & 1e-3/0 \\
H$_2$O & PG-19 & 5e-5/0 & 1e-3/0 \\
xRAG & PG-19 & 5e-5/0 & 5e-5/0 \\
LoCoCo & PG-19 & 0.05/0 & 2e-4/0.1 \\
CCM & PG-19 & 5e-05/0 & 2e-4/0.1 \\
ICAE & PG-19 & 1e-3/0.1 & 1e-3/0.1 \\
Beacon & PG-19 & 0.0002/0.1 & 2e-4/0.1 \\
MemoryLLM & PG-19 & 0.0002/0.1 & 2e-4/0 \\
M+ & PG-19 & 1e-3/0.1 & 5e-5/0.1 \\

\bottomrule\end{tabular}\caption{Final learning rates and weight decay for the nine-system comparison. Task adaptation, suffix access and development histories differ; these are finite configurations rather than an optimized architecture ranking.}\label{tab:nine-final-rates}\end{table*}

\begin{table*}[t]\centering\small
\setlength{\tabcolsep}{5pt}
\begin{tabular}{@{}lrrrr@{}}\toprule
System & Memory trainables & Local trainables & Synthetic updates (M/L) & PG-19 updates (M/L)\\\midrule
StreamingLLM & 21,561,344 & 21,561,344 & 24 / (183+64+512) & 128 / 128\\
H$_2$O & 21,561,344 & 21,561,344 & 24 / (183+64+512) & 128 / 128\\
xRAG & 33,562,624 & 33,559,552 & 128 / 128 & 128 / 128\\
LoCoCo & 11,012,128 & 11,010,048 & 256 / 256 & 256 / 256\\
CCM & 16,842,752 & 16,777,216 & 256 / 256 & 256 / 256\\
ICAE & 218,636,288 & 218,636,288 & 128 / 128 & 64 / 64\\
Beacon & 462,551,040 & 462,551,040 & 128 / 128 & 128 / 128\\
MemoryLLM & 37,882,880 & 37,882,880 & 128 / 128 & 128 / 128\\
M+ & 180,489,216 & 180,489,216 & 128 / 128 & 64 / 64\\
\bottomrule\end{tabular}
\caption{Task-trainable parameters and actual updates of the final fits behind Table~\ref{tab:main-systems}, three seeds per row. M/L denotes memory/local. Counts exclude frozen backbone/compressor parameters and prior pretraining. Synthetic KV has unequal, staged fitting histories; parameter equality alone does not match optimization.}
\label{tab:nine-counts}\end{table*}

For synthetic StreamingLLM and H$_2$O, the model receiving the retained key/value (KV) cache uses 24 ordinary-language-model updates on a 183-update schedule. Its local comparator instead receives 183 ordinary-language-model updates, then 64 target-only updates at $3\times10^{-5}$ and 512 at $10^{-5}$; the last two stages use zero weight decay. Here target-only means that loss is computed only at designated task tokens. These evaluated training histories have unequal budgets. Other rows train on designated target tokens at batch 64, with rates and weight decay in Table~\ref{tab:nine-final-rates}. LoCoCo, CCM and the ordinary 128-update fits use their endpoint checkpoints after the stated setting selection. The displayed ICAE and M+ PG-19 confirmations evaluate update 64 from a 128-update schedule. The archive also retains the earlier settings, and the supporting evidence lists all development candidates and frozen configurations.

The local comparators' PG-19 NLLs range from 4.488 to 8.665 because each system has its own trained comparison model. These models differ in backbone, tokenizer, accessible input, adapted components and training recipe. For example, the displayed ICAE comparator uses learning rate $10^{-3}$ and weight decay 0.1, evaluated at update 64; LoCoCo's uses $2\times10^{-4}$ and weight decay 0.1 for 256 updates. These differences neither identify the cause of the NLL gap nor establish that the comparators are optimally trained. Raw NLL cannot rank architectures across these conditions. We compute utility only between the memory model and local comparator belonging to the same system and task.

\section{Expanded Learning-Rate Search for QA Projectors}\label{app:projector-grid}
All thirty final QASPER projector fits use learning rate $2\times10^{-4}$, selected at the upper edge of the original three-rate development grid. To check whether larger rates improve the comparison conditions more, we run a bounded development search over $2\times10^{-4},8\times10^{-4},3.2\times10^{-3},1.28\times10^{-2},5.12\times10^{-2}$ separately for document, question and constant input. The search reuses seed 491, the same 128 development questions, batch 16, zero weight decay and a 128-update budget. Minimum NLL selects among checkpoints 0,16,32,64,128. The last two rates are specified before their outputs, and no test result selects a rate. This diagnostic does not retrain the ten-seed test models or replace the original twelve-effect family.

\begin{table}[t]\centering\small
\begin{tabular}{@{}lrrr@{}}\toprule
Learning rate & Document & Question & Constant\\\midrule
$2\times10^{-4}$ & 2.1352 & 2.5067 & 2.4693 \\
$8\times10^{-4}$ & {2.1321} & 2.4550 & 2.4462 \\
$3.2\times10^{-3}$ & 2.3597 & {2.4405} & {2.4402} \\
$1.28\times10^{-2}$ & 2.4451 & 2.4540 & 2.4640 \\
$5.12\times10^{-2}$ & 2.5074 & 2.5078 & 2.5582 \\
\bottomrule\end{tabular}
\caption{Minimum development canonical-answer NLL over the fixed checkpoints (lower is better). Every displayed cell selects update 128. These are development scores, not new test estimates.}
\label{tab:projector-expanded-grid}\end{table}

The best tested rates lie inside the expanded grid: $8\times10^{-4}$ for document input and $3.2\times10^{-3}$ for question and constant input. Larger rates worsen development NLL. Finding better rates for all conditions does not establish equal optimization quality or a global optimum. The reported projector test results still use the original training recipe. The matched decoder comparisons train a different component with fixed memory from the public model and do not depend on this projector selection.

\section{Fixed-Software Generation on NQ-open and WebQuestions}\label{app:native-two-datasets}
We test whether the TriviaQA comparison between document, donor and closed-book input extends to two other QA datasets under the same generation settings. The evaluation includes all 3,610 NQ-open and 2,032 WebQuestions items used in the earlier batch-one comparisons. Questions, answer aliases, top-ranked DPR passages \citep{karpukhin-etal-2020-dense}, cached document vectors and three fixed donor assignments remain unchanged. Retrieval files come from the public unsupervised-passage-reranking data distribution; the evidence archive records their download links and hashes. Passage titles remain metadata and are excluded from encoder input. Donors have different passage IDs, but may still contain relevant facts or answer strings.

We use PyTorch 2.1.1, Transformers 4.38.0, RTX A6000, BF16 weights, the released greedy generation procedure, a 100-token cap and its punctuation stopping rule. Each original batch contains four consecutive items, except the final two-item NQ-open batch. Splitting the evaluation across shards preserves these batches. Closed-book input supplies the question without a vector; document input supplies the retrieved passage vector. Donor scores average three separate runs with the fixed donor vectors. We specify the analysis and the priority of answer-inclusion accuracy (Match) before generating these outputs. Reuse of public tests and possible upstream training exposure prevent an unseen-task confirmation claim.

For each dataset, we compare donor input with closed-book input and document input with donor input, using Match, EM and F1. These two contrasts, two datasets and three scores form a separate twelve-effect family. We report ordinary 95\% and Bonferroni-adjusted intervals from 50,000 paired bootstrap draws. Resampling units are connected components of questions sharing a batch or retrieved passage: 760 components for NQ-open and 231 for WebQuestions. The checkpoint and donor bank stay fixed. This family does not replace the earlier six planned contrasts or the fifteen-effect TriviaQA family. Output-length and question-order checks are descriptive.

\begin{table*}[t]\centering\small
\begin{tabular}{@{}llrrrr@{}}\toprule
Dataset & Input & Match & EM & F1 & Mean output tokens\\\midrule
NQ-open & Closed book & 30.2493 & 0.0000 & 7.9622 & 34.1532 \\
NQ-open & Document & 38.6981 & 0.6094 & 15.5835 & 28.1623 \\
NQ-open & Donor 1 & 30.4432 & 0.1939 & 11.8873 & 28.1850 \\
NQ-open & Donor 2 & 30.0554 & 0.0831 & 11.6641 & 28.9119 \\
NQ-open & Donor 3 & 29.6399 & 0.1385 & 11.4745 & 28.8266 \\
NQ-open & Donor mean & 30.0462 & 0.1385 & 11.6753 & 28.6412 \\
\midrule
WebQuestions & Closed book & 35.2854 & 0.0000 & 11.6513 & 28.5950 \\
WebQuestions & Document & 44.4882 & 0.1476 & 17.4504 & 30.5969 \\
WebQuestions & Donor 1 & 39.3701 & 0.1476 & 15.7946 & 28.2972 \\
WebQuestions & Donor 2 & 39.8622 & 0.1476 & 15.4242 & 28.7761 \\
WebQuestions & Donor 3 & 38.5827 & 0.1476 & 15.6862 & 28.1501 \\
WebQuestions & Donor mean & 39.2717 & 0.1476 & 15.6350 & 28.4078 \\
\midrule
\bottomrule\end{tabular}
\caption{Absolute scores from every condition. Strict EM is near its floor on both datasets. Output length retokenizes saved answers and is not an independent mechanism test.}
\label{tab:native-two-absolute}\end{table*}

\begin{table*}[t]\centering\small
\begin{tabular}{@{}llrrcc@{}}\toprule
Dataset & Contrast & Score & Estimate & Ordinary 95\% & Twelve-effect adjusted\\\midrule
NQ-open & Donor $-$ closed & MATCH & $-0.2031$ & $[-1.3444,0.9311]$ & $[-1.8828,1.4350]$ \\
NQ-open & Donor $-$ closed & EM & $+0.1385$ & $[0.0632,0.2307]$ & $[0.0364,0.2820]$ \\
NQ-open & Donor $-$ closed & F1 & $+3.7131$ & $[3.3262,4.1094]$ & $[3.1503,4.2827]$ \\
NQ-open & Document $-$ donor & MATCH & $+8.6519$ & $[7.4687,9.8528]$ & $[6.9336,10.4153]$ \\
NQ-open & Document $-$ donor & EM & $+0.4709$ & $[0.2227,0.7472]$ & $[0.1285,0.8838]$ \\
NQ-open & Document $-$ donor & F1 & $+3.9082$ & $[3.3504,4.4779]$ & $[3.1030,4.7489]$ \\
WebQuestions & Donor $-$ closed & MATCH & $+3.9862$ & $[2.3347,5.6661]$ & $[1.4908,6.5860]$ \\
WebQuestions & Donor $-$ closed & EM & $+0.1476$ & $[0.0389,0.2743]$ & $[0.0000,0.3517]$ \\
WebQuestions & Donor $-$ closed & F1 & $+3.9837$ & $[3.2389,4.6022]$ & $[2.9108,4.8372]$ \\
WebQuestions & Document $-$ donor & MATCH & $+5.2165$ & $[3.7805,6.7188]$ & $[3.0453,7.4653]$ \\
WebQuestions & Document $-$ donor & EM & $+0.0000$ & $[-0.1244,0.1656]$ & $[-0.1766,0.2653]$ \\
WebQuestions & Document $-$ donor & F1 & $+1.8154$ & $[1.1931,2.5716]$ & $[0.9307,2.9417]$ \\
\bottomrule\end{tabular}
\caption{All twelve effects. The WebQuestions donor EM adjusted lower bound equals zero; its positive point estimate does not establish an adjusted positive effect.}
\label{tab:native-two-effects}\end{table*}

\begin{table}[t]\centering\small
\begin{tabular}{@{}llrr@{}}\toprule
Dataset & Input & Reverse & Regroup\\\midrule
NQ-open & Closed & 0 & 228 \\
NQ-open & Document & 0 & 221 \\
NQ-open & Donor 1 & 0 & 216 \\
NQ-open & Donor 2 & 0 & 223 \\
NQ-open & Donor 3 & 0 & 224 \\
WebQuestions & Closed & 0 & 231 \\
WebQuestions & Document & 0 & 218 \\
WebQuestions & Donor 1 & 0 & 218 \\
WebQuestions & Donor 2 & 0 & 217 \\
WebQuestions & Donor 3 & 0 & 223 \\
\bottomrule\end{tabular}
\caption{Changed raw answers among the first 256 questions under within-batch reversal or frozen regrouping; prompts and input tokens remain identical per item. All score changes are retained in the numerical supplement.}
\label{tab:native-two-orders}\end{table}

On WebQuestions, donor input improves Match over closed-book input even after adjustment. On NQ-open, that interval includes zero. Thus the unresolved donor Match gain observed on TriviaQA does not hold across all datasets. Document input improves Match over donor input on both added datasets. Donor F1 gains over closed-book input are positive, while strict EM gains are small or unresolved. Mean donor output is $-5.51$ tokens shorter than closed-book output on NQ-open and $-0.19$ shorter on WebQuestions, but these observations do not establish answer length as the sole cause. Official rescoring and a separately indexed bootstrap reconstruct the saved answers, contrasts and order diagnostics without rerunning model inference.

\section{Additional QASPER Comparisons}\label{app:qasper-details}
The main text retains the comparisons needed to distinguish document-content benefit from gains without current-document facts. The following tables preserve the full absolute scores and the separate decoder-training study.
\begin{table}[t]
\centering\footnotesize
\setlength{\tabcolsep}{4pt}
\begin{tabular}{@{}lrrr@{}}
\toprule
Input / intervention & F1 & EM & Answer NLL \\
\midrule
Document input & 19.98 & 4.29 & 2.0979 \\
Question input & 12.53 & 3.15 & 2.4336 \\
Constant input & 12.54 & 3.13 & 2.4161 \\
Shared zero-vector input & 6.82 & 0.00 & 10.6852 \\
Document, swapped & 7.09 & 1.44 & 3.0507 \\
\midrule
Full-text reference & 29.57 & 0.20 & 7.4681 \\
Closed-book reference & 6.60 & 0.00 & 11.7111 \\
\bottomrule
\end{tabular}
\caption{Gold-answer QASPER with supplied evidence: all 979 test questions. Trained rows average ten training orders; full-text and closed-book rows share the frozen released decoder. F1/EM are percentages; NLL is example-mean answer-token loss. All arms receive the question. Zeroed projector outputs give the same frozen predictor, shown once (Equation~\ref{eq:common-zero}).}
\label{tab:qa-content-absolute}
\end{table}

\begin{table}[t]\centering\footnotesize
\setlength{\tabcolsep}{3pt}
\begin{tabular}{@{}lrrr@{}}\toprule
Training input & F1 & NLL & F1 gain\\\midrule
Correct document & 21.21 & 1.9941 & ---\\
Wrong document & 13.73 & 2.3289 & $7.48\;[5.78,9.17]$\\
Closed book & 13.27 & 2.4044 & $7.94\;[6.14,9.72]$\\
Full text & 62.54 & 0.6520 & $-41.33$ (exploratory)\\
\bottomrule\end{tabular}
\caption{QASPER decoders: 979 test questions and ten training orders. Both document-trained rows use correct vectors at evaluation. F1 gain is the correct-document score minus each row. NLL is mean canonical-answer token loss in nats; lower is better. Gain intervals resample orders and papers, adjusting for six comparisons (Appendix~\ref{app:followup-families}). The full-text advantage has exploratory crossed 95\% bounds $[38.95,43.68]$.}
\label{tab:reader-main}\end{table}

\begin{table}[t]\centering\footnotesize
\setlength{\tabcolsep}{3pt}
\begin{tabular}{@{}llrr@{}}\toprule
Compressed fit & Reference & Ref. F1 & Gain\\\midrule
Projector (19.98) & Frozen closed book & 6.60 & 13.37\\
Projector (19.98) & Question projector & 12.53 & 7.44\\
Decoder (21.21) & Trained closed book & 13.27 & 7.94\\
Decoder (21.21) & Wrong-document fit & 13.73 & 7.48\\
\bottomrule\end{tabular}
\caption{QASPER gains over comparison models. Gain is compressed-model F1 minus reference F1, computed from unrounded means. All rows share questions; only rows training the same component hold the compressed model fixed.}
\label{tab:qa-reference-comparison}\end{table}

\paragraph{The comparison model matters.}
Table~\ref{tab:qa-reference-comparison} compares gains against baselines. The projector and decoder studies train different components; their scores do not describe one fixed compressed model. In the decoder study, evidence text gives 62.54 F1 versus 21.21 from compressed evidence and 13.27 without evidence. The compressed model closes $(21.21-13.27)/(62.54-13.27)=16.1\%$ of the score gap between no evidence and text. The exploratory full-text advantage is 41.33 points.

In the projector study, the ratio is 58.24\%, using compressed F1 of 19.98 and frozen full-text and closed-book scores of 29.57 and 6.60 (Appendix~\ref{app:qa-routing}). Both the compressed model and its baselines change between studies. These ratios describe performance for each training procedure, not the fraction of document information recovered.

\section{Reconciling the QA Comparisons}
\label{app:qa-reconciliation}
This post hoc synthesis reuses saved answers; it adds no training or generation. It was motivated by the apparent disagreement between positive document-content results on QASPER and gains from other questions' passages in released-model QA. Existing aggregate results were known before this synthesis, so it is not a preregistered cross-study test.

\paragraph{Different comparisons answer different questions.}
The QASPER projector comparison changes training input while retaining the projector architecture and frozen LLM. Its document-versus-constant gain asks whether training with current-document information improves answers. The later decoder comparison trains on correct documents, changing other documents, or no documents; both document-trained decoders then receive correct memory at evaluation. Its 21.21 versus 13.73 F1 contrast therefore compares training histories. Neither comparison estimates the effect of giving an already trained model another document at inference. The released-model QA comparison does exactly the latter, contrasting retrieved passages, other-question passages and no document without updating weights. Correct-document superiority and an other-document gain over no document are compatible outcomes.

\paragraph{A within-dataset check of training.}
The earlier fixed-memory decoder study (Appendix~\ref{supp:fixed-reader}) evaluates the public model and ten document-trained decoders on the same 979 questions from 388 papers. Correct and other-paper inputs use identical cached memory bytes before and after training; question order, donor assignments, prompt, and greedy EOS-based generation with a 96-token cap are unchanged. We verified saved-output hashes against completion receipts, matched question/evidence identities and memory hashes, and independently rescored all 22,517 saved answers used here, including the no-document panel below. Frozen scientific source hashes remain unchanged.

\begin{table}[t]\centering\small
\setlength{\tabcolsep}{4pt}
\begin{tabular}{@{}lrrr@{}}\toprule
Decoder & Correct & Other paper & Difference\\\midrule
Public, unadapted & 14.21 & 10.90 & 3.31\\
Document-trained & 21.29 & 7.15 & 14.14\\
\bottomrule\end{tabular}
\caption{F1 on identical QASPER inputs before and after decoder training. The trained row averages ten training orders from the earlier fixed-memory study; it is distinct from the later 21.21-F1 decoder comparison. The increase in the correct-minus-other difference is 10.83 points, with original crossed simultaneous 95\% bounds $[9.07,12.59]$.}
\label{tab:qa-training-sensitivity}
\end{table}

Correct-memory F1 rises by 7.08 points, while other-paper F1 falls by 3.74 points. Their difference, 10.8294, reconstructs the original interaction estimate exactly. Its original simultaneous interval resamples paired training orders and papers and retains that study's comparison family. It is not a newly selected marginal interval. The change demonstrates that decoder adaptation alters sensitivity to the supplied memory without changing the dataset or memory. It does not isolate semantic use from changes in answer format, or establish training as the sole cause of differences between datasets.

\paragraph{Other-paper gains also appear in unadapted QASPER.}
We pair the saved public-model other-paper outputs above with the 979 no-document outputs from the projector study. Recorded checkpoint/tokenizer manifests, question identities, answer aliases and the shared generation implementation agree. The public scores are 6.6047 F1 without a document, 10.8990 with another paper's memory, and 14.2055 with correct memory. The other-paper-minus-no-document difference is $+4.2944$ points. A new, descriptive whole-paper bootstrap gives a marginal 95\% interval $[3.5358,5.0890]$ (20,000 draws, seed 20261013). Each draw samples 388 papers with replacement, keeps all questions within each selected paper, and computes the question-weighted paired difference. This interval is unadjusted and separate from the original study families.

These are separate saved evaluation runs; matching source and checkpoint records do not verify complete runtime equivalence. The no-document prompt also omits the background instruction. Thus the contrast describes the whole input conditions and is not a controlled estimate of document semantics alone. Mean generated lengths are 80.95 tokens without a document, 48.47 with other-paper memory, and 48.90 with correct memory. Length and content change together, so these observations do not establish shorter output as the cause of the F1 increase. They nevertheless show that the descriptive other-document-gain pattern is not unique to the released-model retrieval benchmarks.

\paragraph{Scoring and generation change the interpretation.}
In the final pinned-runtime TriviaQA analysis, no-document, donor and retrieved-passage F1 are 19.38, 24.78 and 29.88. Donors improve F1 by 5.40 points with the original fifteen-comparison bounds $[4.64,6.17]$, while their Match difference is $-0.57\;[-1.50,0.37]$. The retrieved passage exceeds donors on both F1 ($+5.10\;[4.28,5.92]$) and Match ($+9.28\;[8.26,10.31]$). All underlying output-file hashes were checked against the final analysis. These results support passage-specific advantages over the chosen donors, alongside gains over no document that do not require the question's retrieved passage. Donors are not certified answer-free. Neither the F1 differences nor their ratios measure a fraction of information preserved.

QASPER uses annotated scientific-paper evidence, whereas the other benchmarks use retrieved passages and different answer distributions. Its EOS-based generation also differs from the released repository's batch-dependent stopping. The batch-size analysis demonstrates sensitivity within released-model QA, but does not separate output length from padding or numerical changes. The existing experiments identify a within-QASPER training effect and within-benchmark generation/scoring sensitivity. They do not identify the individual causal contributions of dataset, retrieval quality, answer format, or runtime to a cross-study difference. CoQA's short annotated evidence and upstream training exposure likewise prevent treating its high scores as a matched counterexample to the retrieval experiments.

\end{document}